\documentclass{article}

\usepackage{PRIMEarxiv}

\usepackage[utf8]{inputenc} % allow utf-8 input
\usepackage[T1]{fontenc}    % use 8-bit T1 fonts
\usepackage{hyperref}       % hyperlinks
\usepackage{url}            % simple URL typesetting
\usepackage{booktabs}       % professional-quality tables
\usepackage{amsfonts}       % blackboard math symbols
\usepackage{nicefrac}       % compact symbols for 1/2, etc.
\usepackage{microtype}      % microtypography
\usepackage{fancyhdr}       % header
\usepackage{graphicx}       % graphics
\usepackage{amsmath}
\usepackage{amssymb}
\usepackage{amsthm}
\usepackage{subcaption}
\graphicspath{{media/}}
\usepackage{natbib}
\usepackage{xcolor}
\usepackage{cancel}

\newtheorem{theorem}{Theorem}
\newtheorem{lemma}[theorem]{Lemma}

\newtheorem{definition}[theorem]{Definition}

\newtheorem{statement}[theorem]{Statement}

\newcommand{\eg}{\textit{e.g.,}}
\newcommand{\ie}{\textit{i.e.}}

\usepackage{algorithm}
\usepackage{algpseudocode}
\usepackage{graphicx}
\usepackage{subcaption}
\usepackage{tabularx}
\usepackage{multirow}

\newcommand{\mymatrix}[1]{\boldsymbol{#1}}
\newcommand{\myvect}[1]{\boldsymbol{#1}}

\usepackage{amsmath,amsfonts,bm, amsthm}

\def\eqref#1{equation~\ref{#1}}
\def\1{\bm{1}}

\DeclareMathAlphabet{\mathsfit}{\encodingdefault}{\sfdefault}{m}{sl}
\SetMathAlphabet{\mathsfit}{bold}{\encodingdefault}{\sfdefault}{bx}{n}

\title{Shortest-Path Flow Matching with Mixture-Conditioned Bases for OOD Generalization to Unseen Conditions}

\author{
\begin{tabular}{c}
{
Andrea Rubbi\textsuperscript{$\ddagger$,1,4}\quad
Amir Akbarnejad\textsuperscript{$\ddagger$,1,5}\quad
Mohammad Vali Sanian\textsuperscript{$\ddagger$,1,2,6}\quad
Aryan Yazdan Parast\textsuperscript{$\ddagger$,3}
}\\
{
Hesam Asadollahzadeh\textsuperscript{1}\quad
Arian Amani\textsuperscript{1}\quad
Naveed Akhtar\textsuperscript{3}\quad
Sarah Cooper\textsuperscript{1}
}\\
{
Andrew Bassett\textsuperscript{1}\quad
Pietro Li\`o\textsuperscript{4}\quad
Lassi Paavolainen\textsuperscript{2}\quad
Sattar Vakili\textsuperscript{1,7}
}\\
{
Mo Lotfollahi\textsuperscript{1,5,8}
}\\[0.6em]
{
\begin{minipage}{\linewidth}
\centering
\normalfont
\textsuperscript{1}Wellcome Sanger Institute, Cambridge, United Kingdom\\
\normalfont
\textsuperscript{2}Institute for Molecular Medicine Finland (FIMM), University of Helsinki, Helsinki, Finland\\
\normalfont
\textsuperscript{3}School of Computing and Information Systems, The University of Melbourne, Melbourne, Australia\\
\normalfont
\textsuperscript{4}Department of Computer Science and Technology, University of Cambridge, Cambridge, United Kingdom\\
\normalfont
\textsuperscript{5}Cambridge Centre for AI in Medicine, University of Cambridge, Cambridge, United Kingdom\\
\normalfont
\textsuperscript{6}Department of Computer Science, University of Helsinki, Helsinki, Finland\\
\normalfont
\textsuperscript{7}MediaTek Research, Cambridge, United Kingdom\\
\normalfont
\textsuperscript{8}Cambridge Stem Cell Institute, University of Cambridge, Cambridge, United Kingdom
\end{minipage}
}
\end{tabular}
}

\begin{document}

\maketitle

\footnotetext[3]{These authors contributed equally to this work.}

% % keywords can be removed
% \keywords{Flow Matching \and Generative Models \and Out-of-Distribution Generalization \and Gaussian Mixture Models}

\begin{abstract}
Robust generalization under distribution shift remains a key challenge for conditional generative modeling: conditional flow-based methods often fit the training conditions well but fail to extrapolate to unseen ones. We introduce \textbf{SP-FM}, a shortest-path flow-matching framework that improves out-of-distribution (OOD) generalization by conditioning both the base distribution and the flow field on the condition. Specifically, SP-FM learns a condition-dependent base distribution parameterized as a flexible, learnable mixture, together with a condition-dependent vector field trained via shortest-path flow matching. Conditioning the base allows the model to adapt its starting distribution across conditions, enabling smooth interpolation and more reliable extrapolation beyond the observed training range. We provide theoretical insights into the resulting conditional transport and show how mixture-conditioned bases enhance robustness under shift. Empirically, SP-FM is effective across heterogeneous domains, including predicting responses to unseen perturbations in single-cell transcriptomics and modeling treatment effects in high-content microscopy--based drug screening. Overall, SP-FM provides a simple yet effective plug-in strategy for improving conditional generative modeling and OOD generalization across diverse domains.
\end{abstract}

\section{Introduction}
\label{sec:intro}

Conditional generative models have achieved remarkable success across domains ranging from image synthesis to molecular design~\citep{rombach2022highresolutionimagesynthesislatent, classifierfreeguidance}. A common paradigm underlying many state-of-the-art approaches, including diffusion models and flow matching, is to learn a transport map from a simple base distribution to a complex target distribution, conditioned on some input descriptor~\citep{lipman2022flow, atongminibatchot, benamou2000computational}. However, a critical limitation emerges when these models encounter conditions not seen during training: their performance degrades substantially, limiting their utility in real-world applications where exhaustive enumeration of all possible conditions is infeasible.

% This challenge of \emph{out-of-distribution (OOD) generalization}\footnote{In this manuscript the term OOD generalization refers to the ability of a model to take in an unseen descriptor and accurately generate the corresponding population. This departs from the common notion of the term, that is how a distribution shift in a predictor's input instances affects its predictions. Moreover, in our setting the unseen descriptor should still come from the same distribution.} is particularly acute in scientific domains. Consider drug discovery, where one aims to predict cellular responses to novel compounds~\citep{subramanian2017next, srivatsan2020massively}, or genetic screening, where the goal is to predict phenotypic effects of unseen perturbations~\citep{Norman2019, Replogle2022}. In such settings, the space of possible conditions (drugs, genetic perturbations, material compositions) is combinatorially vast, and any practical model must extrapolate beyond its training distribution. Recent work has highlighted that even sophisticated deep learning approaches often fail to outperform simple baselines when evaluated on truly held-out conditions~\citep{ahlmanneltze2025deep}, underscoring the difficulty of this problem.

Out-of-distribution (OOD) generalization is particularly challenging in scientific applications because models are routinely asked to generate responses for \emph{conditions that were never observed during training}. Here, we use \emph{condition} to mean the \emph{intervention or experimental setting} that indexes a data-generating distribution---for example a drug identity, a genetic perturbation, or a specific rotation of a letter---rather than a shift in the input instances of a predictor. Conditional generative models are often brittle in this regime: they may interpolate within the training condition set but degrade sharply on truly held-out conditions, producing overly averaged samples or missing condition-specific effects.\footnote{In this paper, OOD generalization refers to generating from an unseen \emph{condition} (intervention/setting) excluded from training, not to test-time covariate shift in predictor inputs. We assume the unseen condition is drawn from the same overall condition distribution.}

This issue appears across scientific domains. For example, in drug discovery one aims to predict cellular responses to novel compounds~\citep{subramanian2017next, srivatsan2020massively}. Similarly, in genetic screening the goal is to predict phenotypic effects of unseen perturbations~\citep{Norman2019, Replogle2022}. In such settings, the condition space (e.g., drugs, perturbations, compositions and their combinations) is combinatorially large, making exhaustive coverage impossible and forcing practical models to extrapolate beyond the training conditions. Consistent with this, recent benchmarks report that even sophisticated deep learning approaches often fail to outperform simple baselines when evaluated on truly held-out conditions~\citep{ahlmanneltze2025deep}, underscoring the difficulty of robust conditional generation under shift.

Why do current conditional generative models struggle with OOD conditions? We argue that the root cause lies in an overlooked architectural choice: \emph{the base distribution is fixed}. Standard conditional flow matching and diffusion models use a fixed Gaussian $\mathcal{N}(0, I)$ as the source distribution, conditioning only the decoder (velocity field or score network) on the input descriptor~\citep{lipman2022flow, Klein2025, Adduri2025}. This design implicitly treats each condition as an independent estimation problem: the model must learn a separate mapping from the same fixed base to each target distribution. When a novel condition arrives at test time, the model has no mechanism to leverage structural similarities with training conditions; it can only hope that the conditioned decoder generalizes due to, \eg{} the decoder being smooth with respect to its input descriptor.

Recent work has begun to move beyond fixed base distributions, but along directions that differ from our objective. 
\citet{chen2025gmflow} model the \emph{velocity field} as a Gaussian mixture to enable efficient few-step sampling, without learning a condition-dependent base distribution. 
\citet{atanackovic2024meta} propose Meta Flow Matching, which conditions the dynamics on an \emph{observed initial population} encoded by a graph neural network, effectively assuming access to the starting distribution at test time and focusing on transferring dynamics across contexts. 
Other approaches construct application-specific priors to shorten transport paths and improve sampling efficiency~\citep{bunne2023learning}. 
In contrast, SP-FM explicitly \emph{learns} a condition-dependent base distribution jointly with the flow field, with the goal of improving OOD generalization to unseen conditions, and provides theoretical motivation for why conditioning the base can matter.

\begin{figure*}[t]
    \centering
    \begin{subfigure}[b]{0.4\textwidth}
        \centering
        \includegraphics[width=\textwidth]{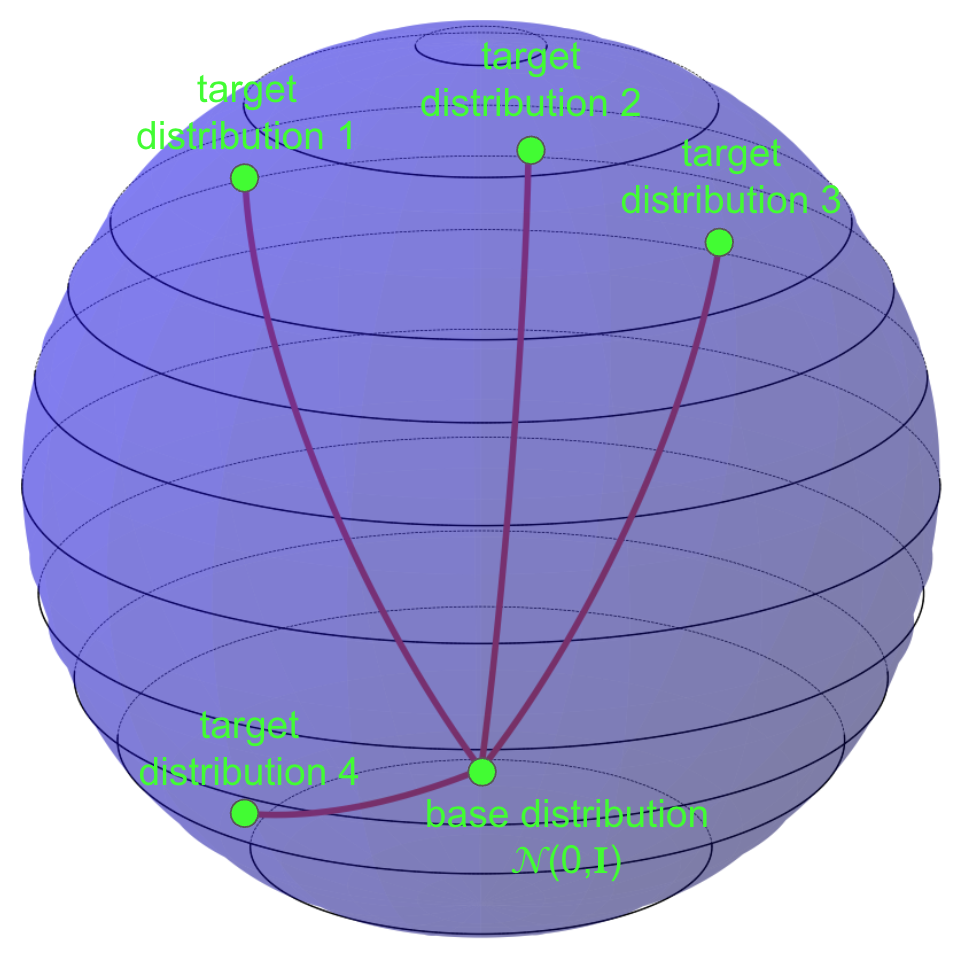}
        \caption{}
        \label{fig:manif1_vanilla}
    \end{subfigure}
    \hfill
    \begin{subfigure}[b]{0.4\textwidth}
        \centering
        \includegraphics[width=\textwidth]{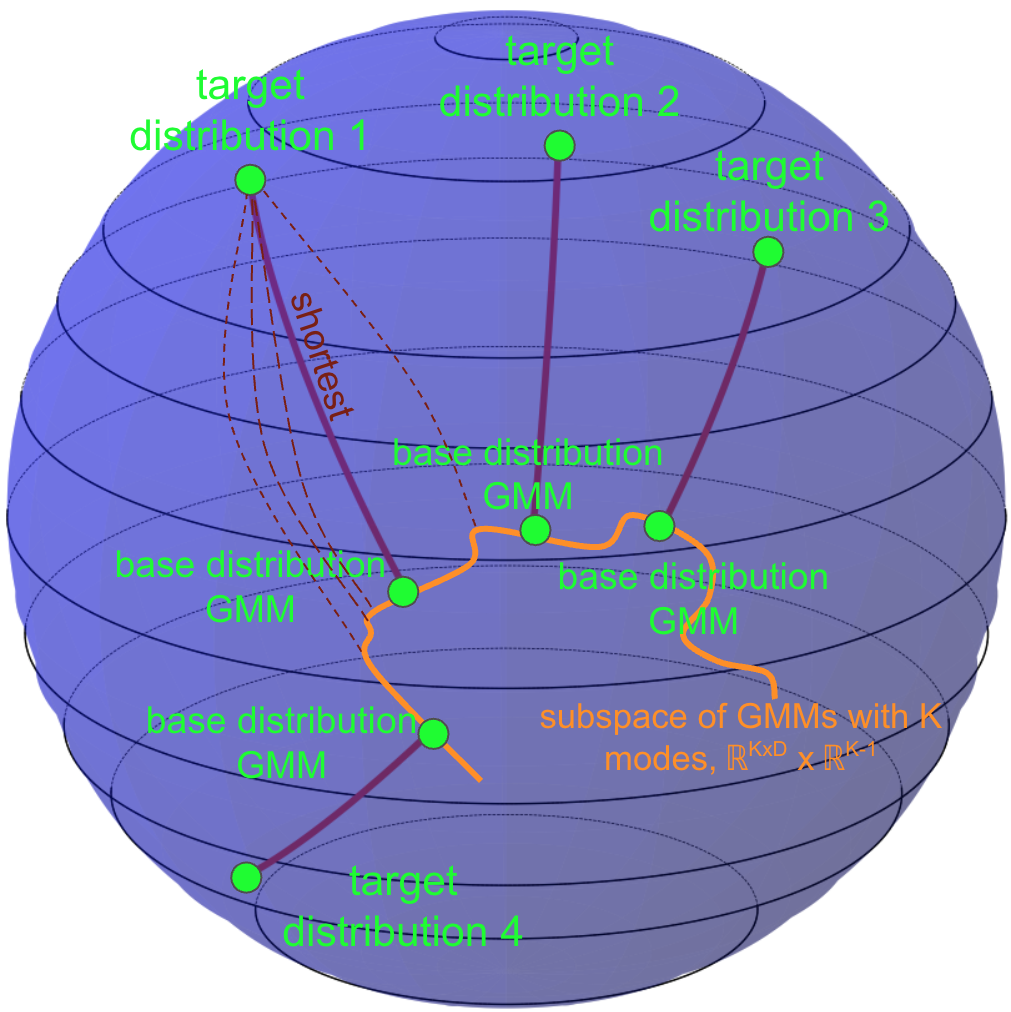}
        \caption{}
        \label{fig:manif2_ptflow}
    \end{subfigure}
        \caption{\textbf{Standard conditional flow matching versus      SP-FM.}
    (a) Conventional approaches fix the base distribution to a standard Gaussian and condition only the velocity field on the descriptor $y$, effectively learning independent flows for each condition.
    (b) SP-FM conditions both the base distribution (as a Gaussian mixture) and the velocity field on the descriptor.
    For each target distribution, the base distribution is selected such that the length of the geodesic connecting the two is minimised. 
    Similar conditions produce similar base distributions, requiring only small flow corrections and enabling smooth extrapolation to unseen conditions.}
    \label{fig:overview}
\end{figure*}

We propose a different perspective: \emph{the base distribution itself should be conditioned on the descriptor}. Our key insight is that when two conditions are similar, whether two rotation angles of images that differ by a small increment, two drugs with related chemical structures~\citep{rogers2010extended}, or two genetic perturbations affecting the same pathway~\citep{Schubert2018PerturbSeq}, their corresponding target distributions are also likely to be similar. By learning a base distribution that reflects this similarity structure, we encode prior knowledge about condition relationships directly into the generative process. The flow then needs only to correct for the residual differences, rather than learning the entire mapping from scratch.

We introduce \textbf{SP-FM}, a conditional flow-matching framework that jointly learns a descriptor-conditioned Gaussian mixture base distribution and a descriptor-conditioned velocity field trained via shortest-path (optimal transport) flow matching. The key architectural change is simple: instead of sampling from a fixed Gaussian, SP-FMuses learned networks to predict condition-specific mixture parameters that define the base distribution. Samples from this adaptive base are then transported to the target via a learned velocity field. Because similar descriptors produce similar base distributions, the model can smoothly interpolate and extrapolate to unseen conditions, a capability that fixed-base approaches fundamentally lack.

We provide theoretical analysis supporting this design. Building on connections between flow matching and optimal transport~\citep{peyre2019computational, benamou2000computational}, we analyze the degrees of freedom in the transport problem as a function of the number of mixture components. We show that when the base is a single Gaussian, the dual optimal transport problem becomes ill-defined, providing a formal explanation for the poor generalization of vanilla conditional flow matching. With multiple mixture components, the problem becomes well-posed, and we derive bounds relating the base complexity, target complexity, and data dimension.

While conditional flow matching is often paired with domain-specific architectures and inductive biases (e.g., in single-cell\cite{Klein2025} and morphology-based perturbation prediction\cite{morphodif,zhang2025cellflux}), our goal is a \emph{general} improvement that can be plugged into existing pipelines. We therefore compare primarily to standard conditional flow-matching baselines under matched capacity and training budgets to avoid confounding gains from task-specific design choices. 
On synthetic rotated-letter benchmarks~\citep{atanackovic2024meta}, SP-FM improves generalization to unseen rotations. On image-based phenotypic assays~\citep{caie2010highcontent, sypetkowski2023rxrx1datasetevaluatingexperimental, bray2016cell}, it reduces distributional distances when predicting morphologies under novel compound treatments. On single-cell transcriptomic datasets~\citep{Norman2019, CPA, Replogle2022}, it accurately predicts gene-expression responses to unseen genetic and chemical perturbations.

\subsection{Contributions} 
Our main contributions are:
\begin{itemize}
    \item We identify the fixed base distribution as a key limiting factor for OOD generalization in conditional flow matching and propose MixFlow, which jointly learns a descriptor-conditioned Gaussian mixture base and a shortest-path velocity field.
    \item We provide theoretical analysis showing that single-Gaussian bases lead to ill-posed inference, while mixture bases with sufficient components yield well-defined problems with bounded estimation error.
    \item We demonstrate consistent improvements over conditional flow matching across eight datasets spanning synthetic, image-based, and transcriptomic domains, establishing SP-FMas a flexible framework for conditional generation with improved OOD generalization.
\end{itemize}

\subsection{Related Work}
\label{sec:related}

\paragraph{Flow Matching and Continuous Normalizing Flows.}
Flow matching has emerged as a scalable framework for training continuous normalizing flows by regressing onto conditional velocity fields along interpolation paths between base and target distributions~\citep{lipman2022flow}. Unlike score-based diffusion models that require simulating stochastic differential equations, flow matching enables simulation-free training and deterministic sampling via ordinary differential equations. Extensions have incorporated optimal transport couplings to straighten trajectories and improve sample efficiency~\citep{atongminibatchot, benamou2000computational}, while others have explored Riemannian geometries and alternative path constructions~\citep{peyre2019computational}. A key limitation shared by these approaches is the reliance on a fixed base distribution---typically a standard Gaussian---which we argue fundamentally limits generalization to unseen conditions.

\paragraph{Conditional Generative Models.}
Conditioning generative models on auxiliary information has enabled applications from class-conditional image synthesis~\citep{classifierfreeguidance, rombach2022highresolutionimagesynthesislatent} to text-to-image generation. In scientific domains, conditional models have been applied to predict cellular responses to perturbations~\citep{CPA, hetzel2022predicting, roohani2024predicting}, model molecular dynamics~\citep{bunne2023learning}, and generate cell morphologies~\citep{palma2025predicting, morphodif, zhang2025cellflux}. Recent flow-based approaches include CellFlow~\citep{Klein2025}, which applies flow matching to single-cell phenotype modeling, and STATE~\citep{Adduri2025}, a transformer-based architecture for context-aware perturbation prediction. These methods condition the velocity field or decoder on perturbation descriptors but maintain a fixed base distribution, treating each condition as an effectively independent estimation problem.
Despite their success in different domains, generative models generalization remains a puzzling phenomenon \cite{closedformFM}. For example, a line of research shows that generative models may learn to merely generative training instances, failing to generate novel samples and risking privacy issues when trained on licenced or sensitive medical data \cite{diffusionGeneralizeWhenFail2Memorize, closedformFM, cureduchampFM}.
Besides the memorisation viewpoint, one can see the lack of generalisation from the lens of score estimation \cite{scorematchingPaper}, where the generative model is shown to learn an independent score estimator for each descriptor in the training set and its corresponding population, as well as for the combined population \cite{classifierfreeguidance}. In this case, the independent score estimators are not expected to generalise to unseen descriptors.

\paragraph{Gaussian Mixtures in Generative Models.}
Gaussian mixture models (GMMs) have a long history as flexible density estimators with universal approximation properties~\citep{nguyen2019approximation, zheng2020universal}. Recent work has revisited GMMs in the context of modern generative models. \citet{chen2025gmflow} propose GMFlow, which models the \emph{velocity distribution} as a Gaussian mixture to capture multimodal flow directions, improving few-step sampling and reducing over-saturation artifacts. \citet{ahamed2025molsnap} use GMMs in latent spaces for molecular generation. These approaches modify how the velocity or score is represented but do not condition the \emph{base distribution} on input descriptors. In contrast, SP-FMlearns a descriptor-conditioned GMM as the source distribution, explicitly encoding condition similarity structure to improve OOD generalization.

\paragraph{Learned and Adaptive Prior Distributions.}
Several works have explored moving beyond fixed Gaussian priors. In the context of variational autoencoders, learned priors have been used to improve expressiveness~\citep{bereket2023modelling}. For flow-based models, \citet{bunne2023learning} learn transport maps via neural optimal transport for single-cell perturbation modeling, though with a focus on paired control-treatment settings rather than OOD generalization. Meta Flow Matching~\citep{atanackovic2024meta} conditions on initial populations by embedding them via graph neural networks, enabling generalization across different starting distributions on the Wasserstein manifold. This approach is complementary to ours: while Meta Flow Matching learns to adapt to varying initial populations, SP-FMlearns to predict appropriate base distributions from condition descriptors, directly targeting the OOD generalization problem.

\paragraph{Out-of-Distribution Generalization.}
OOD generalization remains a central challenge in machine learning, with extensive work on domain adaptation, invariant learning, and distributionally robust optimization. In the context of perturbation biology, recent analyses have shown that deep learning models often fail to generalize beyond training perturbations, sometimes underperforming simple linear baselines~\citep{ahlmanneltze2025deep}. This has motivated the development of methods that explicitly encode biological structure, such as pathway information~\citep{roohani2024predicting} or chemical similarity~\citep{subramanian2017next, Wang2016DrugConnectivity}. SP-FMtakes a complementary approach by encoding similarity structure directly into the generative model architecture through the learned base distribution, providing a general mechanism for OOD generalization that does not require domain-specific inductive biases.
\section{Problem Setup}
\label{sec:problem}

We study population-level generative modelling.  
Let $\{\rho_1,\ldots,\rho_N\}$ denote $N$ probability measures on $\mathbb R^D$, 
each corresponding to a distinct population.  
For each $n\in\{1,\ldots,N\}$ we observe a dataset
\[
\mymatrix{X}_n = \{\myvect{x}_{n,1},\ldots,\myvect{x}_{n,s_n}\} \subset \mathbb R^D,
\]
consisting of $s_n$ samples drawn i.i.d.\ from $\rho_n$.  
Each population is associated with a descriptor $\myvect{y}_n \in \mathcal Y$ that encodes 
metadata about the population (e.g., the rotation angle of a letter, perturbation applied to cells, a molecular embedding of a drug, or cell-type information).  

Given training data $\{(\mymatrix{X}_n,\myvect{y}_n)\}_{n=1}^N$, 
the aim is to learn a conditional generative model that produces samples from $\rho_n$ 
when provided with $\myvect{y}_n$.  
At test time, the model is presented with an unseen descriptor 
$\myvect{y}_{\mathrm{test}}$ and must generate samples approximating the corresponding, 
unknown distribution $\rho_{\mathrm{test}}$.  
This setup captures, for example, predicting the distribution of gene expression states 
under new perturbations, conditioned on information about the perturbation, baseline cells, and biological experimental conditions.

\subsection{Background on Conditional Flow Matching}
\label{sec:cfm}

Flow matching provides a scalable framework for learning conditional generative models 
that map a simple base distribution to a potentially-complex one.  

\paragraph{Flow-based generative models.}
A flow model generates a target distribution $\rho$ by transporting a simple base distribution 
$\mu$ (e.g.\ a standard Gaussian) along the trajectories of an ODE.  
Let $v:\mathbb R^D\times [0,1]\to\mathbb R^D$ be a time-dependent velocity field and consider
\begin{equation}
\frac{d}{dt}\myvect{x}_t \;=\; v(\myvect{x}_t,t), 
\qquad \myvect{x}_0 \sim \mu.
\end{equation}
The measure on $\myvect{x}_t$ is denoted by $\rho_t$, which evolves according to the continuity equation: 
\begin{equation}
\partial_t \rho_t \;=\; -\nabla\cdot(\rho_t v(\cdot,t)),
\qquad \rho_0=\mu.
\end{equation}
With an appropriate choice of $v$, the terminal distribution $\rho_1$ coincides with the target $\rho$.

\paragraph{Conditional flows.}
To capture population variability, we parameterize a conditional velocity field
\[
v_\theta:\mathbb R^D\times [0,1]\times \mathcal Y \to \mathbb R^D,
\]
and define the dynamics
\begin{equation}
\frac{d}{dt}\myvect{x}_t 
\;=\; v_\theta(\myvect{x}_t,t;\myvect{y}),
\qquad \myvect{x}_0 \sim q.
\end{equation}
For a descriptor $\myvect{y}_n$, the terminal distribution of this flow, 
denoted $\rho_{1,\theta}(\cdot\mid \myvect{y}_n)$, should approximate $\rho_n$.  
At test time, for a novel descriptor $\myvect{y}_{\mathrm{test}}$, the model produces 
$\rho_{1,\theta}(\cdot\mid \myvect{y}_{\mathrm{test}})$ as an approximation to $\rho_{\mathrm{test}}$.

\paragraph{Flow matching objective.}
Training $v_{\theta}$ is performed by regression to a teacher velocity field that imposes the boundary conditions (\ie{} the conditions $\rho_0=\mu$ and $\rho_1=\rho$) along with potentially other conditions to, \eg{} regularise the flow model.  
The teacher flow is constructed by integrating over interpolant paths between the base $\mu$ and the empirical populations $\rho_n$, while the training is shown to be feasible by individual interpolant paths \cite{lipman2022flow}.
Specifically, let $(\myvect{x}_0,\myvect{x}_1)\sim \mu\otimes \rho_n$, \ie{} \ $\myvect{x}_0\sim \mu$ and 
$\myvect{x}_1\sim \rho_n$ independently, and consider the linear interpolant
\[
\myvect{x}_t \;=\; (1-t)\myvect{x}_0 + t\myvect{x}_1,\qquad t\in[0,1].
\]
The individual conditional teacher interpolant path is defined as
\begin{equation}
u(\myvect{x}_t,t\; | \myvect{x}_0, \myvect{x}_1) \;=\; \myvect{x}_1 - \myvect{x}_0.
\end{equation}
More advanced constructions use Wasserstein geodesic interpolants 
\citep{benamou2000computational,peyre2019computational, atongminibatchot}, 
which provide velocity fields aligned with optimal transport.

The conditional flow matching loss is then
\begin{equation}
\label{eq:cfm-loss}
\mathcal L(\theta) \;=\; 
\frac{1}{N}\sum_{n=1}^N 
\mathbb E_{t\sim \mathrm{Unif}[0,1]}
\mathbb E_{(\myvect{x}_0,\myvect{x}_1)\sim \mu\otimes \rho_n}
\left\|
v_\theta\!\big((1-t)\myvect{x}_0+t\myvect{x}_1,t;\myvect{y}_n\big)
- \big(\myvect{x}_1-\myvect{x}_0\big)
\right\|^2.
\end{equation}
Minimizing~\eqref{eq:cfm-loss} learns a parametric family of velocity fields that transports the base distribution $\mu$ to each $\rho_n$, while sharing parameters across populations through conditioning on $\myvect{y}_n$.

\section{Mixture-Conditioned Flow Matching}
\label{sec:pertflow}

Standard conditional flow matching fixes the base distribution to $\mathcal{N}(\mathbf{0}_D,\sigma^2\mathbf{I}_{D\times D})$ and conditions only the decoder, effectively yielding independent estimators for each population and limiting generalization (See, Fig.~\ref{fig:manif1_vanilla}). SP-FM instead conditions both the base distribution and the flow module on perturbations and covariates (Fig.~\ref{fig:manif2_ptflow}).  

Let $I$ denote the number of mixture components. Plausible base distributions are Gaussian mixtures $\mu_{\mathrm{GMM}(\mymatrix{\Theta}, \myvect{p})}$ with $\mymatrix{\Theta} \in \mathbb{R}^{I\times D}$ specifying mode locations, $\myvect{p}\in\mathbb{S}^{I-1}$ specifying mixture weights, and fixed component variance $\sigma^2$. The family of such mixtures forms the restricted subspace illustrated in Fig.~\ref{fig:manif2_ptflow} in orange.  

Given populations $\{\rho_1,\ldots,\rho_N\}$ with descriptors $\{\myvect{y}_1,\ldots,\myvect{y}_N\}$, each $\rho_n$ is projected to its closest element $\mu_{\mathrm{GMM}(\mymatrix{\Theta}^{(n)}, \myvect{p}^{(n)})}$ in this subspace. The optimal transport to $\rho_n$ from its projection defines a velocity field $v(.,t;\myvect{y}_n)$. These velocity fields form geodesics on the 2-Wasserstein manifold of probability measures, which are shown by the red curves in Fig.~\ref{fig:manif2_ptflow}.
Predictors $h_{\mymatrix{\Theta}}:\mathcal{Y}\to\mathbb{R}^{I\times D}$ and $h_{\myvect{p}}:\mathcal{Y}\to\mathbb{S}^{I-1}$ are trained so that $h_{\mymatrix{\Theta}}(\myvect{y}_n)\approx \mymatrix{\Theta}^{(n)}$ and $h_{\myvect{p}}(\myvect{y}_n)\approx \myvect{p}^{(n)}$. At test time, the predicted base $\mu_{\mathrm{GMM}(h_{\mymatrix{\Theta}}(\myvect{y}_{test}),h_{\myvect{p}}(\myvect{y}_{test}))}$ is evolved by $v(.,t;\myvect{y}_{test})$ to approximate $\rho_{test}$.

\subsection{Training objectives}
SP-FM is trained with two complementary losses.  

\noindent\textbf{Objective 1.} Align the learned velocity field $v_\theta(.,t;\myvect{y}_n)$ with the OT path between $\rho_n$ and its projected base:
\begin{equation}\label{eq:loss_V_matches_OTpath}
    \mathcal{L}_{\mathrm{OT}} =
    \mathbb{E}\Bigg[
    \sum_{s=1}^{S}
    \big\|
        v\!\left(t,(1-t)\myvect{x}^{(0)}_s+t\myvect{x}_{\tau(s)},\myvect{y}_n\right) - (\myvect{x}_{\tau(s)}-\myvect{x}^{(0)}_s)
    \big\|^2
    \Bigg],
\end{equation}
where $\{\myvect{x}_s\}\sim\rho_n$, $\{\myvect{x}^{(0)}_s\}\sim \mu_{GMM(h_{\mymatrix{\Theta}}(\myvect{y}_n),h_{\myvect{p}}(\myvect{y}_n))}$, $t\sim\mathrm{Unif}(0,1)$, and $\tau$ denotes mini-batch OT pairing~\citep{atongminibatchot}.  

\noindent\textbf{Objective 2.} Encourage each projection $\mu_{GMM(h_{\mymatrix{\Theta}}(\myvect{y}_n),h_{\myvect{p}}(\myvect{y}_n))}$ to remain close to $\rho_n$ by minimizing geodesic length:
\begin{equation}\label{eq:loss_geod_len}
    \mathcal{L}_{\mathrm{geo}} =
    \mathbb{E}\Bigg[
    \sum_{s=1}^{S}
    \big\|\myvect{x}_{\tau(s)} - \myvect{x}^{(0)}_s\big\|^2
    \Bigg],
\end{equation}
with the Gumbel–softmax trick ensuring differentiability with respect to mixture weights and mode locations. 

The training procedure is elaborated in Alg.~\ref{alg:pertflow}, where the training alternates between updating the flow model in line 19 and the base distributions in line 22. 
In Alg.~\ref{alg:pertflow} the training planner $P$ divides the training into three parts: (i) warm-up period, where only the flow model is updated, (ii) alternating period, during which after a certain number of updates to the flow model, a single update is made to the base distributions. (iii) cool-down period, where the base distributions are kept fixed and only the flow model is trained. Notably, during the warm-up and alternating periods the dropout layers of $H_{\mymatrix{\Theta}}$ and $H_{\myvect{p}}$ are enabled, while in the cool-down period they are disabled.

\subsection{Inference}
At test time, given a novel descriptor $\myvect{y}_{test}$, SP-FM predicts the base distribution $\mu_{\mathrm{GMM}(h_{\mymatrix{\Theta}}(\myvect{y}_{test}),h_{\myvect{p}}(\myvect{y}_{test}))}$ and evolves it with $v(.,t;\myvect{y}_{test}),\;t\in[0,1]$ to generate samples approximating $\rho_{test}$.

%%%

\begin{algorithm}[t]
\caption{Training SP-FM with mixture-conditioned flow matching}
\label{alg:pertflow}
\begin{algorithmic}[1]
\Require Datasets $\{(\mymatrix{X}_n,\myvect{y}_n)\}_{n=1}^N$, number of base GMM modes $I$, fixed GMM variances $\sigma^2$, training planner $P$.
\State Initialize parameters: $params(v(.,.,.))$ (velocity field), $params(h_{\mymatrix{\Theta}}(.))$ (mode position predictor), and
$params(h_{\myvect{p}}(.))$ (mode probability predictor), learning rate $\ell$.
\State $cnt\_epoch \leftarrow 0$
\State $cnt\_iteration \leftarrow 0$
\For{each training epoch}
    \While{the epoch hasn't finished}
        \State $mode\_train,\;flag\_settoeval\_H \leftarrow P.next(cnt\_epoch,\;cnt\_iteration)$
        \State Sample a population $(\mymatrix{X}_n,\myvect{y}_n)$
        \State $\{\myvect{x}_s\}_{s=1}^{S} \leftarrow$ $S$ samples from the selected population
        \State $\{\myvect{x}^{(0)}_s\}_{s=1}^S \leftarrow S\; \text{Samples from }\mu_{GMM(h_{\mymatrix{\Theta}}(\myvect{y}_n), h_{\myvect{p}}(\myvect{y}_n) )}$
        \State $\tau \leftarrow$ mini-batch discrete OT pairing between $\{\myvect{x}_s\}_{s=1}^{S}$ and $\{\myvect{x}^{(0)}_s\}_{s=1}^{S}$
        \If{$flag\_settoeval\_H$}
            \State $H_{\mymatrix{\Theta}}.eval()$; \;$H_{\mymatrix{p}}.eval()$ \Comment{\ie{} disable dropout}
        \Else
            \State $H_{\mymatrix{\Theta}}.train()$; \;$H_{\mymatrix{p}}.train()$ \Comment{\ie{} enable dropout}
        \EndIf
        \If{$mode\_train \text{\; equals 'velocity field'}$}
            \State $t \leftarrow unif(0,1)$
            \State $\mathcal{L}_{OT} \leftarrow \text{compute according to Eq.~\ref{eq:loss_V_matches_OTpath}}$
            \State $params(v(.,.,.)) \leftarrow params(v(.,.,.)) - \ell\cdot\nabla \mathcal{L}_{OT}$
        \ElsIf{$mode\_train \text{\; equals 'base distribution'}$}
            \State $\mathcal{L}_{geo} \leftarrow \text{compute according to Eq.~\ref{eq:loss_geod_len}}$
            \State $params(H_{\mymatrix{\Theta}}, H_{\myvect{p}}) \leftarrow params(H_{\mymatrix{\Theta}}, H_{\myvect{p}}) - \ell\cdot\nabla \mathcal{L}_{geo}$
        \Else
            \State "print(ERROR: unknown value for $mode\_train$)"
        \EndIf
        \State $cnt\_iteration \leftarrow cnt\_iteration + 1$
    \EndWhile
    \State $cnt\_epoch \leftarrow cnt\_epoch + 1$
    
\EndFor
\end{algorithmic}
\end{algorithm}

\section{Generalization Analysis}
\label{sec:generalization}

The Wasserstein manifold of probability measures is infinite-dimensional, hindering the adoption of theoretical generalization guarantees for finite-dimensional spaces.  
To circumvent this, we assume that the target distribution $\rho$ is itself a Gaussian mixture model (GMM) with $J$ components,
\[
\rho = \rho_{GMM(\mymatrix{\Gamma}, \myvect{q})},
\]
where $\mymatrix{\Gamma}\in\mathbb{R}^{J\times D}$ are fixed mode locations, $\myvect{q}\in\mathbb{S}^{J-1}$ are mixture weights, and all components share a fixed variance. This setting still captures highly general distributions: as $J \to \infty$ and $\mymatrix{\Gamma}$ forms a sufficiently fine grid covering $\mathbb{R}^D$, the family $GMM(\mymatrix{\Gamma},\myvect{q})$ is dense in the space of probability measures under the Wasserstein metric \citep{nguyen2019approximation,zheng2020universal}. Thus, assuming $\rho$ is GMM-induced is a broad and flexible modelling assumption. In this formulation, prediction reduces to estimating the mixture weights $\myvect{q}\in\mathbb{S}^{J-1}$ from the population descriptor $\myvect{y}$.

A naive solution is to directly learn a predictor mapping $\myvect{y}\mapsto \myvect{q}$. However, this approach suffers from poor generalization as $J$ grows, since the dimensionality of $\myvect{q}$ directly affects the generalization. MixFlow instead takes a structured approach: rather than predicting $\myvect{q}$ directly, it (i) projects the target distribution onto the subspace of GMMs with $I$ components (possibly, $I \ll J$), and (ii) learns the velocity field that transports this projection to the target distribution.

Besides the aforementioned assumptions, in the theoretical analysis we assume that this 2-step algorithm uses mixture Wasserstein distance \citep{mixturewassdistOriginalPaper, mixturewassdistTopicModels} and the flow model induced by it, as opposed to the commonly used 2-Wasserstein distance. In the latter case, the pushforward function of the flow model can map its input points to any arbitrary point in $\mathbb{R}^D$. While in the former case, a discrete optimal transport problem is solved between the source and target GMMs to determine, intuitively, how the modes of the source GMM should be moved or split to create the target GMM, and that fully determines the flow.
In this sense, the latter metric and flow model are a generalisation of the former.
Nonetheless, this analysis highlights the importance of conditioning the base distribution on population descriptor $\myvect{y}$, which we experimentally validate in in Sec.~\ref{sec:ablations}.

In this section we only summarize the key theoretical findings. 
Details are provided in Appendix~\ref{sec:socalledsection5:v2:mixflow}.

\begin{statement}[Summary of theoretical findings]\label{statement:maintext:theories:summary}
Recall that $I, J, D \in \mathbb{N}$ are respectively the number of modes in the base and target GMM distributions, and the data dimension. 
\begin{enumerate}\itemsep0pt
\item[§1] In the testing phase, the velocity field $\mymatrix{V}\in \mathbb{R}^{I\times J}$ is identified up to $J - ID$ degrees of freedom. In particular, if $I \ge \lceil J/D \rceil$, then $\mymatrix{V}$ is uniquely identified. 
\item[§2] If $I=1$, the problem reduces to predicting the $J$ mixture weights $\myvect{q}$ directly from $\myvect{y}$, corresponding to the common baseline of fixing the base distribution to $\mathcal{N}(\myvect{0}_D, \mymatrix{I}_{D\times D})$ and explaining its poor generalization.

% \item[§3] Since the entries of $\mymatrix{V}$ are bounded in $[0,1]$, the residual $J-ID$ degrees of freedom in §1 imply a maximum estimation error of \emph{[to be completed]}.

\end{enumerate}
\end{statement}

Essentially, Statement~\ref{statement:maintext:theories:summary} suggests that increasing the number $I$ of modes of the base distribution by one, reduces degrees of freedom of the flow model by the data dimension $D$, thereby facilitating the prediction and identification of the flow model in the testing phase.
% Moreover, in Statement~\ref{statement:maintext:theories:summary}, §2 discourages the adoption of a unimodal GMM, \ie{} a multi-variate Gaussian distribution, as the base distribution. 
% This analytical result, give us a rule of thumb that if we use $J$ as an indicator of the complexity of learnable target distributions, it scales with $I$ the number of modes in the based multiplied by data dimesnion (?)

\section{Experiments}\label{sec:experiments}

We evaluate SP-FM against Vanilla Conditional Flow Matching (Vanilla CFM) on a set of synthetic and experimental datasets.  Our goal is to assess SP-FM's generalizability and its robustness across multiple conditions, datasets, and hyperparameters. We seek empirical evidence of the need for the Gaussian Mixture Model (GMM) to describe the base distribution and its impact on the distributional distance between the inferred final distributions and the empirical ones. 

%%%%%%%%%%%%%%%%%%%%%%%%%%%%%%%%%%%%%%%%
% Sythetic Data
%%%%%%%%%%%%%%%%%%%%%%%%%%%%%%%%%%%%%%%%

\subsection{SP-FM on Synthetic Data}
Following Meta Flow Matching~\citep{atanackovic2024meta}, we create a synthetic benchmark of
populations $\{(\rho_i,\myvect{y}_i)\}$ where each condition $i$ corresponds to a target population
$\rho_i = p_1(x_1\mid i)$ supported on a rendered letter silhouette. For every alphabet character we
generate 20 rotations; each (letter, rotation) pair defines one condition (\ie{} one descriptor $\myvect{y}_i$) and its corresponding population. The descriptor $\myvect{y}_i$ is the concatenation of a one-hot letter code and a normalized rotation
value. The target population $\rho_i$ is formed by sampling uniformly from the foreground pixels of
the silhouette. To ensure that each condition corresponds to a full population rather than a single sample, we
render multiple copies of the same silhouette using different fixed RGB colours. These colour
variations are used only to generate several samples per condition and are not included in the
descriptor $\myvect{y}_i$.

For SP-FM, we construct a mixture-conditioned base for each population. We set the number of
base modes $I$ equal to the number of conditions and assign one Gaussian component per condition.
In our synthetic setup, we do not use $h_{\mymatrix{\Theta}}(\myvect{y}_i)$ to predict the component
means; instead, the GMM mode locations are implemented as free learnable parameters, each
initialized from a single rendered sample of its corresponding condition. The mixture weights are
still predicted by the MLP $h_{\myvect{p}}(\myvect{y}_i)$. 
%\akbar{Is it the case in the latest code?}
The velocity field
$v(\cdot,t;\myvect{y}_i)$ is then trained using the flow-matching objectives in
Eqs.~\ref{eq:loss_V_matches_OTpath}--\ref{eq:loss_geod_len} to transport the resulting
$\mu_{\mathrm{GMM}(\mymatrix{\Theta}^{(i)},h_{\myvect{p}}(\myvect{y}_i))}$ toward $\rho_i$. As a baseline, conditional flow matching (CFM) fixes the base to $\mathcal{N}(\mathbf{0}_D,\mathbf{I}_{D\times D})$ and samples $x_0\sim\mathcal{N}(\mathbf{0},\mathbf{I})$ at generation time.

\begin{figure*}[t]
\centering
\includegraphics[width=\textwidth]{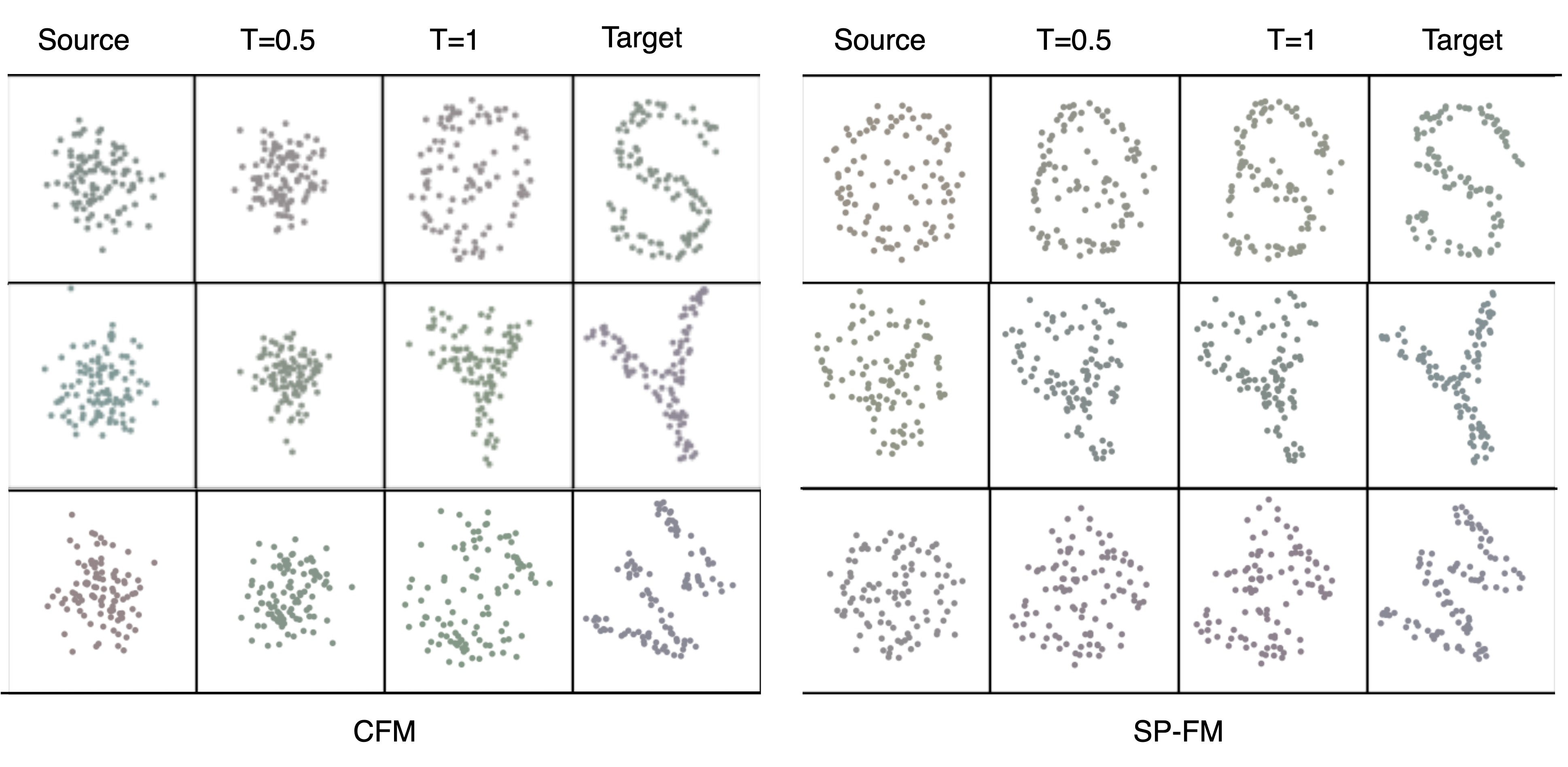}
\caption{
%\akbar{4th and 8th columns should be called Target rather than Source.}
Qualitative comparison of Conditional Flow Matching (CFM) and SP-FM on the synthetic
letter–rotation benchmark. Each block shows samples at different integration times ($t=0$, $t=0.5$,
$t=1$) and the corresponding target distribution. SP-FM produces smoother trajectories
and more faithful reconstructions on unseen rotations, while CFM exhibits drift
and shape distortion.
}
\label{fig:synthetic_qual}
\end{figure*}

We train SP-FM and all baselines on a synthetic dataset of 6 letters with $R=20$ rotations per letter, and study rotation interpolation generalization. Specifically, we generate $R$ rotation angles uniformly spaced in $[0,2\pi]$ and split them by index: the training set contains the 10 even-indexed rotations, while the held-out set contains the 10 odd-indexed rotations. We form the validation set by selecting a single \texttt{val\_letter} and evaluating that letter only on the held-out (odd-indexed) rotations; all letters (including \texttt{val\_letter}) are present in training but only at the even-indexed rotations. In the main paper we report results with $\texttt{val\_letter}=\texttt{S}$, and evaluate generalization on two unseen test letters, \texttt{W} and \texttt{Y}, each measured on their held-out (odd-indexed) rotations. To ensure conclusions are not tied to a particular character geometry, we repeat the same protocol with different choices of validation letter; results for $\texttt{val\_letter}=\texttt{H}$ are reported in the appendix. Results are reported in Table~\ref{tab:synthetic_combined_reformatted}, with qualitative samples in Figure~\ref{fig:synthetic_qual}.

We compare SP-FM against three baselines: (i) \textbf{CFM}, which uses a fixed isotropic Gaussian base; (ii) a \textbf{conditional GMM} baseline that learns a mixture model by expectation-maximization (EM) algorithm with condition-dependent mixture weights $\pi(\cdot\mid \myvect{y})$ (implemented as a learned lookup table over conditions) and shared component parameters, but does not learn a transport map; and (iii) \textbf{SP-FM (K=1)}, an ablation where SP-FM is restricted to a single mixture component, reducing the base distribution to a unimodal Gaussian (learnable mean) and testing whether improved performance can be explained by a single adaptive mode alone. These baselines isolate the contributions of the two key ingredients in SP-FM: learning a condition-adaptive multi-modal base distribution and learning a conditional transport field.

For each setting, we select the best checkpoint based on the validation W2 distance and report all metrics at this checkpoint. Table~\ref{tab:synthetic_combined_reformatted} shows that SP-FM consistently outperforms all baselines across validation on \texttt{S} and testing on \texttt{W} and \texttt{Y}, demonstrating the effectiveness of mixture-conditioned flow matching for rotation-interpolation generalization. Importantly, the relative behavior of the baselines also highlights the contribution of each component: SP-FM (K=1) underperforms, indicating that a unimodal (single-mode) base is insufficient even when it is learnable, while the conditional GMM (EM) baseline underperforms SP-FM, indicating that adapting the base distribution alone (via $\pi(\cdot\mid\myvect{y})$) is not enough without learning the subsequent transport field. Consistent with the training objective that explicitly encourages the predicted base to stay close to the target (Eq.~\ref{eq:loss_geod_len}), the Source vs. Target metrics are improved under SP-FM, providing direct empirical evidence that the learned source distribution is already similar to the target before transport. This behavior is also visible qualitatively in Figure~\ref{fig:synthetic_qual}: SP-FM starts from a source distribution that resembles the target shape and produces smoother, more faithful trajectories, whereas CFM exhibits larger mismatch at $t=0$ and more pronounced distortions along the path. Implementation details and hyperparameters for SP-FM and all baselines are provided in Appendix~\ref{sec:app:synthetic_details}.

%\akbar{I found the setup description slightly ambiguous. (i) how was the validation data created? (ii) are they "two experiments"? (iii) in the test phase, how many unseen rotations of S and H are used?}

\begin{table*}[t]
\centering
\caption{Synthetic letter--rotation benchmark. Models are trained on even-indexed rotations and validated on \texttt{S} using held-out odd-indexed rotations; results are reported for \texttt{S} and unseen rotations on test letters \texttt{W} and \texttt{Y}. We report both Generated vs. Target and \emph{Source vs. Target} distances. The conditional GMM baseline is trained via EM (no validation checkpoint), and since it has no flow model its Source vs. Target equals Generated vs. Target, so only the latter is reported. }
\label{tab:synthetic_combined_reformatted}
\resizebox{\linewidth}{!}{
\begin{tabular}{llcccccc}
\toprule
\multirow{2}{*}{Dataset} & \multirow{2}{*}{Model} 
& \multicolumn{3}{c}{Generated vs. Target} 
& \multicolumn{3}{c}{Source vs. Target} \\
\cmidrule(lr){3-5} \cmidrule(lr){6-8}
& & MMD $\downarrow$ & W1 $\downarrow$ & W2 $\downarrow$
  & MMD $\downarrow$ & W1 $\downarrow$ & W2 $\downarrow$ \\
\midrule

\multirow{3}{*}{Letter S (Val)}
  & CFM      & 0.01627 $\pm$ 0.00069 & 0.5874 $\pm$ 0.0199 & 0.7071 $\pm$ 0.0266
             & 0.08249 $\pm$ 0.00138 & 1.0671 $\pm$ 0.0076 & 1.1898 $\pm$ 0.0069 \\
  & SP-FM  & \textbf{0.00986 $\pm$ 0.00014} & \textbf{0.4740 $\pm$ 0.0028} & \textbf{0.5744 $\pm$ 0.0006}
             & \textbf{0.01986 $\pm$ 0.00008} & \textbf{0.6464 $\pm$ 0.0014} & \textbf{0.7388 $\pm$ 0.0005} \\
  & SP-FM (I=1)  & -- & -- & --
             & -- & -- & -- \\

\midrule
\multirow{3}{*}{Letter W (Test)}
  & CFM      & 0.01809 $\pm$ 0.00081 & 0.6180 $\pm$ 0.0141 & 0.7283 $\pm$ 0.0152
             & 0.06088 $\pm$ 0.00049 & 0.8972 $\pm$ 0.0062 & 1.0151 $\pm$ 0.0071 \\
  
  & SP-FM (I=60)  & \textbf{0.01279 $\pm$ 0.00131} & \textbf{0.5323 $\pm$ 0.0203} & \textbf{0.6310 $\pm$ 0.0223}
             & \textbf{0.01825 $\pm$ 0.00062} & \textbf{0.6101 $\pm$ 0.0067} & \textbf{0.7126 $\pm$ 0.0061} \\
  &GMM (I=600)      & 0.08931 $\pm$ 0.00387 & 1.1311 $\pm$ 0.0299 & 1.3202 $\pm$ 0.0375
           & -- & -- & -- \\
  & SP-FM (I=1)  & -- & -- & --
             & -- & -- & -- \\

\midrule
\multirow{3}{*}{Letter Y (Test)}
  & CFM      & 0.02333 $\pm$ 0.00197 & 0.6278 $\pm$ 0.0160 & 0.7415 $\pm$ 0.0176
             & 0.06020 $\pm$ 0.00128 & 1.0614 $\pm$ 0.0048 & 1.2410 $\pm$ 0.0054 \\
  & SP-FM (I=60)  & \textbf{0.01799 $\pm$ 0.00015} & \textbf{0.5433 $\pm$ 0.0044} & \textbf{0.6403 $\pm$ 0.0064}
             & \textbf{0.04079 $\pm$ 0.00225} & \textbf{0.9336 $\pm$ 0.0297} & \textbf{1.0861 $\pm$ 0.0312} \\

  &GMM (I=600)        & 0.07423 $\pm$ 0.00528 & 0.8986 $\pm$ 0.0317 & 1.0508 $\pm$ 0.0417
           & -- & -- & -- \\

  & SP-FM (I=1)  & -- & -- & --
             & -- & -- & -- \\

\midrule

\end{tabular}%
}
\end{table*}

%%%%%%%%%%%%%%%%%%%%%%%%%%%%%%%%%%%%%%%%
% Single-Cell Data
%%%%%%%%%%%%%%%%%%%%%%%%%%%%%%%%%%%%%%%%

\subsection{Using SP-FM to predict single-cell responses for unseen interventions}
Single-cell perturbation screenings are an invaluable tool for new treatment discovery; they consist in testing the effects of genetic or chemical perturbations on cells in vitro. Each cell is exposed to a specific condition, and its gene expression is measured to assess whether the perturbation is associated with a given phenotypic change.
We evaluate whether SP-FM can predict cellular responses to perturbations never observed before. 

Models are trained on a subset of conditions and tested on held-out ones, defining an \emph{out-of-distribution (OOD) generalisation} problem. Table~\ref{tab:datasets} provides an overview of the datasets used in our experiments. Notably, Combo-Sciplex \citep{CPA} and Norman \citep{Norman2019} datasets also contain combinations of two perturbations. For each dataset, we generate four random splits, holding out 70\% of the perturbations from the training set. 
Given the larger number of cells available in the iAstrocytes and Replogle-Nadig \citep{Replogle2022} datasets, we guarantee high reliability of our results by filtering for conditions with at least 300 and 500 cells, respectively. The iAstrocytes data has been produced in-house in Sanger Institute, and it is not currently publicly available.

We report Wasserstein-1, Wasserstein-2,  MMD, and Energy distances (ED). These metrics assess whether predicted distributions accurately capture the global geometry of observed cell states, beyond low-order statistics. On all tested single-cell perturbation datasets, SP-FM achieves substantially stronger distributional alignment than Vanilla CFM (Table \ref{tab:single_cell_results}), showing how SP-FM better generalises when generating previously unseen conditions, across both genetic and chemical perturbation modalities.

\begin{table}[t]
\centering
\caption{Single-cell perturbation datasets used to evaluate SP-FM.}
\label{tab:datasets}
\renewcommand{\arraystretch}{1.1}
\setlength{\tabcolsep}{5pt}
\begin{tabularx}{\linewidth}{lccc}
\toprule
Dataset & Cells & Perturbations & Cell type \\
\midrule
Norman~\citep{Norman2019}{} & $\sim$90k & 277 CRISPR$^{*}$ & K562 \\
Combo-Sciplex~\citep{CPA} & $\sim$70k & 32 compounds$^{*}$ & A549 \\
iAstrocytes  & $\sim$150k & 148 CRISPR & iAstrocytes \\ 
Replogle--Nadig~\citep{Replogle2022} & $\sim$250k & 67 CRISPRi & Jurkat \\
BBBC021~\citep{caie2010highcontent} & 97{,}504 images & 35 compounds & MCF7 \\
RxRx1 \citep{sypetkowski2023rxrx1datasetevaluatingexperimental} & 170{,}193 images & 1066 siRNA & U2OS \\
\midrule
\end{tabularx}
\small $^{*}$single or paired perturbations.
\end{table}

%We benchmark \textbf{SP-FM} against both established models and simple baselines to assess its ability to predict cellular responses. As strong deep learning baselines, we consider \textbf{STATE} \citep{Adduri2025}, a transformer-based architecture that integrates contextual embeddings for perturbation prediction, and \textbf{CellFlow} \citep{Klein2025}, a flow-matching model designed to capture continuous cell state transitions. Both models are trained and optimised with their recommended hyperparameters on our dataset splits. Our model similar to \textbf{CellFlow} is trained on PCA representation of gene expression data (see lis of hyperparamters for all experiments in 

%To contextualise performance, we additionally implement linear baselines. A low-rank linear regression (LRLR) model estimates perturbation effects as additive shifts relative to control cells, while a naïve baseline returns the average control expression profile. These simple models establish a lower bound on performance and test whether complex architectures capture perturbation-specific effects beyond trivial population-level shifts, a limitation highlighted by \citet{ahlmanneltze2025deep}.  

\begin{table*}[t]
\centering
\caption{Cell-type perturbation benchmarks comparing SP-FM and CFM. Mean $\pm$ std over four runs. Lower is better.}
\label{tab:single_cell_results}
\begin{tabular}{llcccc}
\toprule
Dataset & Model & MMD $\downarrow$ & W1 $\downarrow$ & W2 $\downarrow$ & ED $\downarrow$ \\
\midrule

\multirow{2}{*}{Norman}
  & CFM  & 0.199 $\pm$ 0.009 & 30.449 $\pm$ 0.241 & 467.8 $\pm$ 8.1 & 16.2 $\pm$ 0.3\\
  & SP-FM & \textbf{0.192 $\pm$ 0.004} & \textbf{30.084 $\pm$ 0.284} & \textbf{456.3 $\pm$ 9.1} & \textbf{15.9 $\pm$ 0.3} \\
\midrule

\multirow{2}{*}{Combo-SciPlex}
  & CFM & 0.141 $\pm$ 0.005 & 26.007 $\pm$ 0.464 & 344.8 $\pm$ 12.6 & 9.0 $\pm$ 0.3\\
  & SP-FM & \textbf{0.129 $\pm$ 0.006} & \textbf{25.640 $\pm$ 0.180} & \textbf{335.3 $\pm$ 4.8} & \textbf{8.5 $\pm$ 0.2}\\
\midrule

\multirow{2}{*}{iAstrocytes}
  & CFM      & 0.163 $\pm$ 0.003 & 27.121 $\pm$ 0.110 & 369.7 $\pm$ 3.0 & 13.2 $\pm$ 0.1\\
  & SP-FM & \textbf{0.135 $\pm$ 0.004} & \textbf{27.061 $\pm$ 0.233} & \textbf{367.6 $\pm$ 6.5} & \textbf{ 12.6 $\pm$ 0.1}\\
\midrule

\multirow{2}{*}{Replogle (Jurkat)}
  & CFM      & 0.116 $\pm$ 0.006 & 52.919 $\pm$ 0.200 & 1408.3 $\pm$ 10.6 & 28.0 $\pm$ 0.3\\
  & SP-FM & \textbf{0.095 $\pm$ 0.003} & \textbf{52.868 $\pm$ 0.215} & \textbf{1404.9 $\pm$ 11.5} & \textbf{27.3 $\pm$ 0.4}\\
\bottomrule
\end{tabular}
\end{table*}

%%%%%%%%%%%%%%%%%%%%%%%%%%%%%%%%%%%%%%%%
% Image Data
%%%%%%%%%%%%%%%%%%%%%%%%%%%%%%%%%%%%%%%%

\subsection{SP-FM predicts unseen intervention responses in image-based assays}

To extend beyond transcriptomic data, we evaluate SP-FM in an image-based drug discovery setting. High-content imaging provides rich phenotypic readouts, but exhaustive compound testing is infeasible. We therefore train on a subset of compounds and predict cell morphologies for unseen compounds using only their chemical representations at test time. 

We benchmark on BBBC021~\cite{caie2010highcontent} and RxRx1~\cite{sypetkowski2023rxrx1datasetevaluatingexperimental}, comparing SP-FM against a conditional flow matching baseline (CFM). Table~\ref{tab:bbbc021_rxrx1_benchmarks} shows that SP-FM achieves consistently lower distance metrics (MMD, W1, W2, ED), indicating more faithful recovery of unseen phenotypic responses. We split BBBC021 following prior work \citep{palma2025predicting}: training uses 25 compounds and testing 9. For RxRx1, we train on 853 siRNAs and test on 213. 

A key design choice is to generate instances in \emph{feature space} rather than pixel space. Images are encoded with DINOv2 \citep{oquab2024dinov2learningrobustvisual} into $768$-D embeddings, which aligns with semantically meaningful morphology-level distances \citep{oquab2024dinov2learningrobustvisual, zhang2018unreasonableeffectivenessdeepfeatures}. This strategy improves efficiency, imposes a semantic inductive bias, and enables geometry-aware evaluation. For fairness, CFM outputs are also scored in the same DINOv2 feature space. Results are provided in Table 4.
These substantial gains indicate that SP-FM more faithfully recovers unseen compound phenotypes, highlighting its potential to reduce experimental burden through reliable \textit{in~silico} prediction.

% SP-FM generalizes more strongly than CFM, reducing W1 distances by $\sim$73\% (BBBC021) and $\sim$74\% (RxRx1), W2 by $\sim$92\% and $\sim$93\%, MMD by $\sim$82\% and $\sim$88\%, and ED by $\sim$87\% and $\sim$92\%. 
 
\begin{table*}[t]
\centering
\caption{BBBC021 and RxRx1 benchmarks comparing SP-FM and CFM on phenotypic perturbations. Lower is better; values are the mean over three runs.}
\label{tab:bbbc021_rxrx1_benchmarks}
\begin{tabular}{llcccc}
\toprule
Dataset & Model & MMD $\downarrow$ & W1 $\downarrow$ & W2 $\downarrow$ & ED $\downarrow$ \\
\midrule
\multirow{2}{*}{BBBC021}
  & CFM      & 0.2504 $\pm$ 0.001 & 108.4093 $\pm$ 0.291 & 5903.8647 $\pm$ 32.33 & 18.9197 $\pm$ 0.088 \\
  & SP-FM & \textbf{0.0439} $\pm$ 0.002 & \textbf{29.4752} $\pm$ 0.733 & \textbf{442.9421} $\pm$ 24.11 & \textbf{2.4036} $\pm$ 0.087 \\
\midrule
\multirow{2}{*}{RxRx1}
  & CFM      & 0.2527 $\pm$ 0.001 & 118.7191 $\pm$ 1.23 & 7077.8330 $\pm$ 43.21 & 20.5810 $\pm$ 0.31 \\
  & SP-FM & \textbf{0.0307} $\pm$ 0.001 & \textbf{30.8660} $\pm$ 0.12 & \textbf{483.2796} $\pm$ 4.69 & \textbf{1.6731} $\pm$ 0.034 \\
\bottomrule
\end{tabular}
\end{table*}

\section{Ablations}\label{sec:ablations}

We conduct ablation studies to understand the key design choices in SP-FM. All experiments are performed on both transcriptomic (iAstrocytes) and morphological (BBBC021) datasets to ensure findings generalize across modalities.

\begin{figure*}[t]
  \centering
  \begin{subfigure}[t]{0.7\linewidth}
    \centering
    \includegraphics[width=\linewidth]{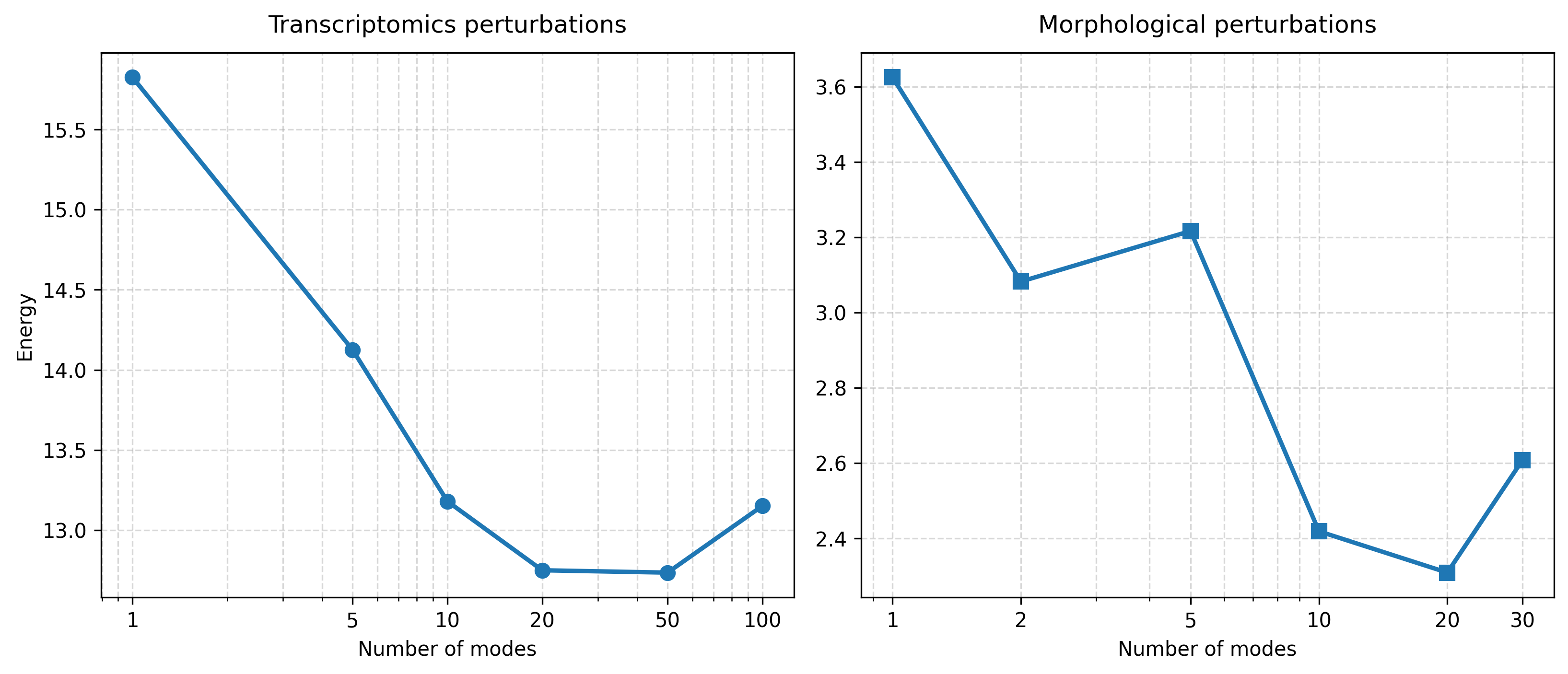}
    \caption{Energy distance versus number of GMM modes on transcriptomic (left) and morphological (right) datasets. Performance improves with more modes up to a dataset-dependent optimum.}
    \label{fig:ablation:modes}
  \end{subfigure}
  \vspace{0.6em}
  \begin{subfigure}[t]{0.7\linewidth}
    \centering
    \includegraphics[width=\linewidth]{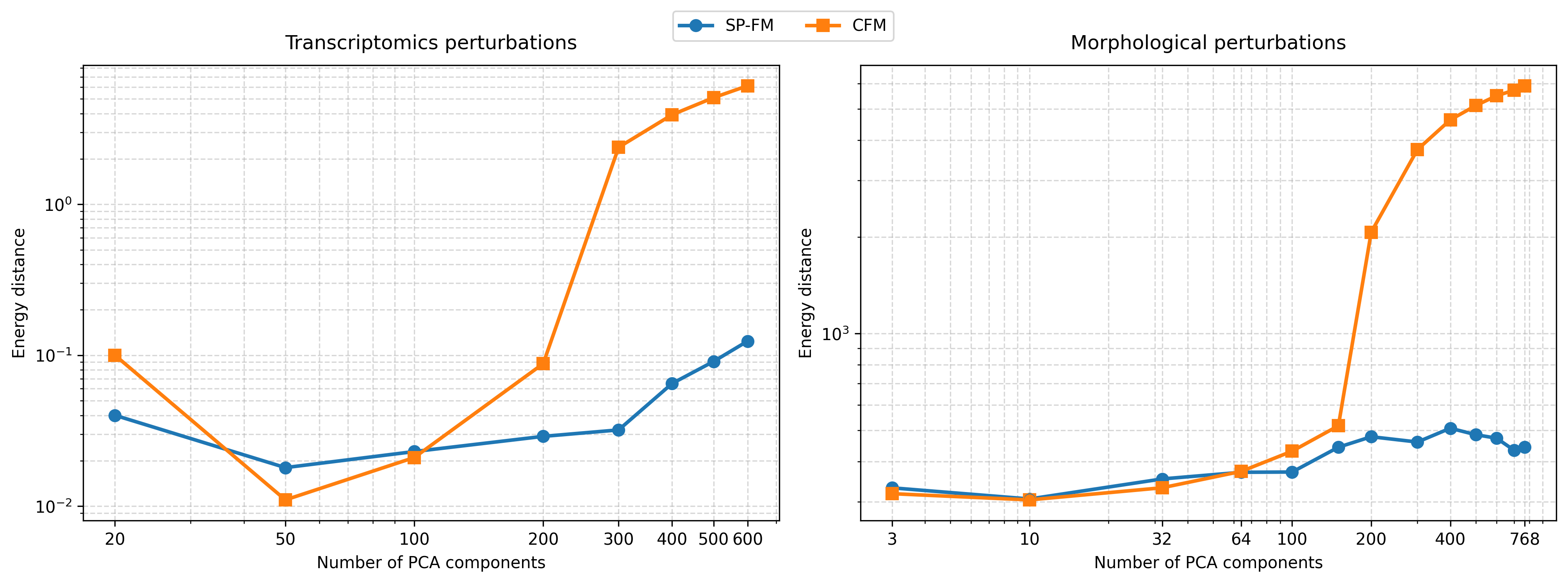}
    \caption{Energy distance versus PCA dimensionality for SP-FM and CFM. SP-FM maintains stable performance while CFM degrades in higher dimensions.}
    \label{fig:ablation:pcs}
  \end{subfigure}
  \caption{\textbf{Ablation studies} on transcriptomic (Combo-SciPlex) and morphological (BBBC021) datasets. (a)~Effect of the number of GMM modes on generation quality. (b)~Comparison of SP-FM and CFM robustness across increasing data dimensionality.}
  \label{fig:ablation}
\end{figure*}

\paragraph{Number of mixture components.}
A central design choice in SP-FM is the number of modes $I$ in the Gaussian mixture base distribution. We vary $I$ (the number of modes) and measure the energy distance between generated and ground-truth distributions on held-out perturbations. As shown in Figure~\ref{fig:ablation:modes}, increasing the number of modes consistently improves performance up to a point: energy distance decreases as $I$ grows, reflecting the increased expressiveness of the base distribution. However, beyond a dataset-dependent threshold, performance plateaus or slightly degrades. We attribute this to optimization difficulty: with too many modes, the model must learn to allocate probability mass across components that may not be necessary, increasing the effective number of parameters without proportional benefit. Notably, the optimal $I$ differs between datasets (transcriptomic data benefits from more modes than morphological data) suggesting that the appropriate complexity of the base distribution should reflect the heterogeneity of the target populations. These findings align with our theoretical analysis in Section~\ref{sec:generalization}: increasing $I$ reduces the degrees of freedom in the transport problem by $D$ per mode, improving identifiability (Statement~1, \S2).

\paragraph{Robustness to data dimensionality.}
A known challenge for flow-based models is scaling to high-dimensional spaces, where the fixed Gaussian base distribution $\mathcal{N}(0, I)$ becomes increasingly mismatched to structured target distributions. We investigate this by training both SP-FM and CFM on PCA-reduced representations with increasing numbers of components. Figure~\ref{fig:ablation:pcs} shows that CFM's performance degrades substantially as dimensionality increases: the energy distance grows by over an order of magnitude between 50 and 500 PCA components. In contrast, SP-FM maintains stable performance across the full range of dimensionalities tested. This robustness arises because the learned GMM base distribution adapts to the structure of each condition, reducing the effective transport distance even in high dimensions. From the perspective of Section~\ref{sec:generalization}, higher $D$ increases the constraint count ($ID + I$), which for fixed $I$ reduces the residual degrees of freedom $J - ID$, explaining why SP-FM's advantage becomes more pronounced in higher dimensions.

These ablations confirm two key properties of SP-FM: (i) the number of mixture components provides a useful complexity knob that should be tuned to dataset heterogeneity, and (ii) conditioning the base distribution yields substantial robustness benefits in high-dimensional settings where standard CFM struggles.

\section{Limitations and Future Work}

While SP-FM demonstrates strong performance across diverse perturbation modalities, several limitations remain. The framework depends on the availability of high-quality perturbation descriptors such as gene embeddings or chemical fingerprints; incomplete or noisy metadata may hinder predictive accuracy. Our evaluation focuses primarily on single perturbations and small combinations, whereas scaling to higher-order combinations, dose–response landscapes, and time-course dynamics remains an open challenge. Moreover, although we provide theoretical justification for out-of-distribution generalisation, the analysis assumes smooth transport fields that may not fully capture heterogeneous tissues or disease states.  

Future work should address these challenges by incorporating richer multimodal descriptors, extending the model to capture interaction effects and temporal responses, and scaling training to very large datasets. In particular, applying SP-FM to resources such as the JUMP- consortium dataset for imaging assays \citep{chandrasekaran2024three} and large single-cell perturbation compendia \citep{zhang2025tahoe} could enable the development of true zero-shot models that leverage perturbation similarity at scale. Finally, extending the framework to translational settings such as patient-derived cells or organoids will be critical for realising its potential impact in therapeutic design.

\section*{Acknowledgments}

This work is supported by Open Targets (Drug2Cell) and by the Chan Zuckerberg Initiative DAF Single-Cell Biology Data Insights program.

\section*{Conflict of interest}
M.L. has equity interests in Relation Therapeutics, is a scientific co-founder and part-time employee of AIVIVO, and serves on the scientific advisory board of Novo Nordisk.

\section*{Code availability}
Code is publicly available on github: \href{https://github.com/Lotfollahi-lab/MixFlow}{https://github.com/Lotfollahi-lab/MixFlow}.

%Bibliography
\bibliographystyle{plainnat}  
\bibliography{main}  

\newpage

\appendix

\section{Generalisation analysis, details and proofs}\label{sec:socalledsection5:v2:mixflow}
As outlined in Sec.~\ref{sec:generalization}, to circumvent the infinite-dimensional Wasserstein manifold of measures and to enable the adoption of theoretical generalization guarantees for finite-dimensional spaces, we assume that the target distribution $\rho$ is itself a Gaussian mixture model (GMM) with $J$ components, fixed mode locations that cover $\mathbb{R}^D$ as $J\rightarrow \infty$, and varying mixture weights. Moreover, recall that $\mu$ and $\rho$ denote the base and target distributions, respectively. 

\begin{definition}[Mixture Wasserstein Distance]
Let $\mu_{GMM(\myvect{\Theta}, \myvect{p})}$ and $\rho_{GMM(\myvect{\Gamma}, \myvect{q})}$ be two measures induced by their corresponding GMMs with $I$ and $J$ modes, mode positions $\myvect{\Theta}\in \mathbb{R}^{I\times D}$ and $\myvect{\Gamma}\in \mathbb{R}^{J\times D}$, and mode weights $\myvect{p} \in \mathbb{S}^{I-1}$ and $\myvect{q} \in \mathbb{S}^{J-1}$. The mixture Wasserstein distance between $\mu_{GMM(\myvect{\Theta}, \myvect{p})}$ and $\mu_{GMM(\myvect{\Gamma}, \myvect{q})}$ is defined as follows:

\begin{align}
    \mathcal{M}\mathcal{W}^2_2(\mu, \rho) &= \underset{\myvect{V}\in \mathbb{R}^{I\times J}}{inf} \; \sum_{i=1}^{I} \sum_{j=1}^J ||\myvect{\Theta}_i-\myvect{\Gamma}_j||_2^2 \; v_{ij} \; p_i\label{eq:kantorovich_objective}\\
    \text{s.t.}&\; \sum_{j=1}^J v_{ij} =1 \;\;, \;\; \forall i\in \{1,...,I\}\\
    &\;\;q_j = \sum_{i=1}^I v_{ij}\; p_i\;\;, \;\;\forall j \in \{1,....,J\} \;\;\; \text{(boundary condition)}\\
    &\;\;v_{ij} \geq 0 \;\; , \;\; \forall i\in \{1,...,I\},\; \forall j\in \{1,...,J\}
\end{align}
This is equivalent to discrete optimal transport between the GMMs' modes with the transportation cost determined by their mode locations. Intuitively, $v_{ij}$ is the portion of mass of the $i$-th mode in the 1st GMM $\mu$ which is moved to the $j$-th mode in the 2nd GMM $\rho$.
\end{definition}

\begin{definition}[Mixture Wasserstein Flow]
The matrix $\myvect{V}^{I\times J}$ in Eq. \ref{eq:kantorovich_objective} induces a time-dependant flow ${v:\mathbb{R}^D \times [0,1] \rightarrow \mathbb{R}^D}$ referred to as mixture Wasserstein flow.
\end{definition}

\begin{lemma}[Dual of Mixture Wasserstein Distance]
The dual form of the constrained optimisation of Eq. \ref{eq:kantorovich_objective} is as follows:
\begin{align}
    &\underset{\myvect{z}\in \mathbb{R}^{I + J}}{sup} \; \sum_{i=1}^{I}z_i + \sum_{j=1}^J q_{j} \; z_{j+I}\label{eq:kantorovich_dualform}\\
    \text{s.t.}&\; z_i+ p_i\cdot z_{j+I}\; \leq p_i ||\myvect{\Theta}_i-\myvect{\Gamma}_j||_2^2\;\;, \;\; \forall i\in \{1,...,I\}, \; j\in \{1,...,J\} \label{eq:kantorovich_dual_IJconstraints}\\
    &\;\; \{z_1, z_2, ..., z_{I+J}\} \;\;\; \text{unrestricted in sign , i.e. $\in \mathbb{R}$}
\end{align}
\end{lemma}

\begin{definition}[Subset sum condition]\label{def:subsetsum}
In the discrete optimal transport problem of Eq. \ref{eq:kantorovich_objective}, the marginal measures $[p_1,...,p_I]$ and $[q_1,...,q_J]$ are said to meet the subset sum condition if for any proper non-empty subsets of them - denoted by $\{p_{i1},...,p_{i_{I'}}\}$ and $\{q_{j1},...,q_{j_{J'}}\}$- we have that 
\begin{equation}
   ( p_{i1} + ... + p_{i_{I'}}) \neq 
    (q_{j1}+...+q_{j_{J'}}).
\end{equation}
\end{definition}

\begin{statement}\label{statement:subsetsum_ifandonlyif_dualandprimal}
In the discrete optimal transport problem of Eq. \ref{eq:kantorovich_objective} assume that the marginal distributions meet the subset sum condition. Then we have that:  
\begin{enumerate}\itemsep0pt
    \item Any optimal solution of Eq. \ref{eq:kantorovich_objective}, $\mymatrix{V}^*\in \mathbb{R}^{I\times J}$, has exactly $I+J-1$ non-zero elements. 
    \item A dual constraint in Eq. \ref{eq:kantorovich_dual_IJconstraints} is binding if and only if the corresponding $\mymatrix{V}^*[i,j] = v^*_{ij}$ is non-zero.
    \item The dual problem of Eq. \ref{eq:kantorovich_dualform} has a unique optimum $\myvect{z}^* \in \mathbb{R}^{I+J}$.
\end{enumerate}
% \akbar{I've checked the statement above in google search, AI mode, too.}
\end{statement}

The importance of Statement~\ref{statement:subsetsum_ifandonlyif_dualandprimal} is that - as we will prove in Statement~\ref{statement:subsetsum:almostsurely} - for any $\mu$ and $\rho$ the subset sum condition, and hence 1, 2, and 3 explained above hold almost surely. Therefore, we can generally assume 1, 2, and 3 in Statement~\ref{statement:subsetsum_ifandonlyif_dualandprimal} hold for any $\mu$ and $\rho$.

\begin{statement}\label{statement:Vstargt0_unique}
In the discrete optimal transport problem of Eq. \ref{eq:kantorovich_objective} assume that the marginal distributions meet the subset sum condition.
Moreover, assume that $\myvect{\Theta}$, $\myvect{\Gamma}$, $\myvect{p}$, and $[z^*_1, ..., z^*_I]$ in the dual form are known and fixed. Then $\myvect{V}^{*}_{\ge 0}$ is uniquely determined, where $\myvect{V}^* \in \mathbb{R}^{I\times J}$ is an optimal solution for Eq. \ref{eq:kantorovich_objective}, and $\myvect{V}^*_{\ge 0} \in \mathbb{R}^{I\times J}$ is a matrix where the element in its $i$-th row and $j$-th column is 1 if $v^*_{ij} > 0$ and is 0 otherwise.

\textbf{Proof}: The constraint of Eq. \ref{eq:kantorovich_dualform} rewrites as
\begin{equation}
    z_{j+I} \le \Big[||\myvect{\Theta}_i-\myvect{\Gamma}_j||_2^2 - \frac{z_i}{p_i}\Big] 
\end{equation}
To maximise the objective of Eq. \ref{eq:kantorovich_dualform}, the variables $z_{j+I}$ should take their maximum possible value. Therefore we have that

\begin{equation}\label{eq:kantrovich_dual_complementary}
    z^*_{j+I} = \underset{1\le i \le I}{max}\Big[||\myvect{\Theta}_i-\myvect{\Gamma}_j||_2^2 - \frac{z^*_i}{p_i}\Big].
\end{equation}
In other words, given $[z^*_1, ..., z^*_I]$ the above equation uniquely specifies the values for $[z^*_{I+1}, ..., z^*_{I+J}]$ and regardless of the probability values of the target distribution $\myvect{q} \in \mathbb{R}^J$. 
According to Statement \ref{statement:subsetsum_ifandonlyif_dualandprimal}
, in Eq. \ref{eq:kantrovich_dual_complementary} equality holds if and only if $v^*_{ij} > 0$, and this determines $\myvect{V}^*_{\ge 0}$. $\blacksquare$
\end{statement}

The importance of Statement~\ref{statement:Vstargt0_unique} is that, intuitively, having predicted the closest GMM to $\rho$ (i.e. $\myvect{\Theta}$ and $\myvect{p}$), having predicted the first $I$ optimal dual values $[z^*_1,..,z^*_I]\in \mathbb{R}^I$, and having fixed the GMM modes that cover the feature space (i.e. $\myvect{\Gamma}$), $\myvect{V}^*_{\ge 0}$ is uniquely identified.
This is important, because to obtain a predictor $\mymatrix{V}^*_{\ge0} \approx h_{\mymatrix{V}^*_{\ge0}}(\myvect{y})$ where $h_{\mymatrix{V}^*_{\ge0}}:\mathbb{R}^{dim(\myvect{y})}\rightarrow  \{0,1\}^{I\times J}$, one can instead train the following three predictors, and $h_{\mymatrix{V}^*_{\ge0}}(.)$ will be uniquely determined:
\begin{enumerate}\itemsep0pt
    \item $\mymatrix{\Theta}\approx h_{\mymatrix{\Theta}}(\myvect{y})$, where $h_{\mymatrix{\Theta}}:\mathbb{R}^{dim(\myvect{y})}\rightarrow \mathbb{R}^{I\times D}$.
    \item $\myvect{p}\approx h_{\myvect{p}}(\myvect{y})$, where $h_{\myvect{p}}:\mathbb{R}^{dim(\myvect{y})}\rightarrow \mathbb{S}^{I-1}$.
    \item $[z^*_1, ..., z^*_I] \approx h_z(\myvect{y})$, where $h_{z}:\mathbb{R}^{dim(\myvect{y})} \rightarrow \mathbb{R}^I$
\end{enumerate}

\begin{statement}\label{statement:subsetsum:almostsurely}
    Let $\mu_{GMM(\myvect{\Theta}, \myvect{p})}$ and $\rho_{GMM(\myvect{\Gamma}, \myvect{q})}$ be two measures induced by their corresponding GMMs. Then the marginal distributions $\myvect{p}$ and $\myvect{q}$ meet the subset sum condition almost surely. Therefore, Statements \ref{statement:subsetsum_ifandonlyif_dualandprimal} and \ref{statement:Vstargt0_unique} hold almost surely.

\textbf{Proof}: There are $(2^I-2)\times(2^J-2)$ pairs of proper non-empty subsets from the marginal measures $[p_1,...,p_I]$ and $[q_1,...,q_J]$. Each such a pair  
like $\{p_{i1},...,p_{i_{I'}}\}$ and $\{q_{j1},...,q_{j_{J'}}\}$ specify the following subspace of $\mathbb{S}^{I-1}\times \mathbb{S}^{J-1}$ where the subset sum condition is violated:
\begin{equation}
    p_{i1}+...+p_{i_{I'}} = q_{j1}+...+q_{j_{J'}}
\end{equation}
Each such subspace has zero measure, so the union of these $(2^I-2)\times(2^J-2)$ subspaces has also zero measure. $\blacksquare$
\end{statement}

\begin{definition}[Projection to subspace of GMMs with $I$ modes]
Let $\rho_{GMM(\mymatrix{\Gamma}, \myvect{q})}$ be a measure induced by a GMM with modes $\mymatrix{\Gamma} \in \mathbb{R}^{J\times D}$ and mode probabilities $\myvect{q} \in \mathbb{S}^{J-1}$. The goal is to find the closest measure to $\rho_{GMM(\mymatrix{\Gamma}, \myvect{q})}$ which is induced by a GMM with $I$ modes, where $I<<J$ and in terms of mixture Wasserstein distance. This GMM $\mu_{GMM(\mymatrix{\Theta}*, \myvect{p}*)}$ is found by:
\begin{equation}\label{eq:project_toImodeGMMs}
    \mymatrix{\Theta}^*, \myvect{p}^* = \underset{\substack{\mymatrix{\Theta}\in\mathbb{R}^{I\times D} \\ \myvect{p}\in \mathbb{R}^I }}{argmin}
    \;\; \mathcal{M}\mathcal{W}^2_2(\mu_{GMM(\myvect{\Theta}, \myvect{p})}, \rho_{GMM(\myvect{\Gamma}, \myvect{q})})
\end{equation}
% \akbar{According to google search (AI mode), the above objective function is highly non-convex, and is usually solved by gradient-based optimisation.}
\end{definition}

Let's assume we have the 3 predictors $h_{\mymatrix{\Theta}}(.)$, $h_{\myvect{p}}(.)$, and $h_{z}(.)$ explained above, that take in the population descriptor $\myvect{y}$ and predict the projection of the unseen $\rho$ to the space of GMMs (\ie{} $\mymatrix{\Theta}^*$ and $\myvect{p}^*$) as well the $I$ dual solution values $[z^*_1, ..., z^*_I]$. Given these, according to Statement~\ref{statement:Vstargt0_unique} we can uniquely determine $I+J-1$ elements of $\myvect{V}^* \in \mathbb{R}^{I\times J}$ which are non-zero, and the rest are uniquely set to zero. Recall that the goal is to predict the flow $\myvect{V}^* \in \mathbb{R}^{I\times J}$, because when it is applied to the measure $\mu_{GMM(\mymatrix{\Theta}*, \myvect{p}*)}$ it gives us the prediction for $\rho$.
Having identified non-zero elements of $\myvect{V}^*$, Statements~\ref{statement:barycentrecondition} and \ref{statement:pcondition} impose linear constraints on the identified elements, thereby improving the idenfication and prediction of $\myvect{V}^*$.

\begin{statement}[Barycentre condition]\label{statement:barycentrecondition} 
For the matrix $\mymatrix{\Theta}^* \in \mathbb{R}^{I\times D}$ in Eq. \ref{eq:project_toImodeGMMs} we have that
\begin{equation}\label{eq:barycentercondition}
    \myvect{\Theta}^*_i = \sum_{j=1}^J v^*_{ij} \myvect{\Gamma}_j.
\end{equation}
\end{statement}
% \akbar{This definitely holds according to google search (AI mode). But I'm slightly doubtful about my proof below}.\\
\textbf{Proof}: We compute the derivative of Eq. \ref{eq:kantorovich_objective} with respect to $\myvect{\Theta}_i$ and set it to zero:
\begin{align}
    &\frac{\partial}{\partial \myvect{\Theta}_i} \sum_{j=1}^J ||\myvect{\Theta}_i-\myvect{\Gamma}_j||_2^2 \; v^*_{ij} \; p_i =\\
    &\frac{\partial}{\partial \myvect{\Theta}_i} \sum_{j=1}^J \big[
        \myvect{\Theta}_i^T\myvect{\Theta}_i
        - 2 \myvect{\Theta}_i^T \myvect{\Gamma}_j
    \big] \; v^*_{ij} \; p_i =\\
    &\sum_{j=1}^J [2\myvect{\Theta}_i - 2\myvect{\Gamma}_j] v^*_{ij} p_i =\\
    & 2p_i \; [\myvect{\Theta}_i \cancelto{1}{\sum_{j=1}^J v^*_{ij}}- \sum_{j=1}^Jv^*_{ij}\myvect{\Gamma}_j] = 0\\
    \Rightarrow & \myvect{\Theta}_i^* = \sum_{j=1}^J v^*_{ij} \myvect{\Gamma}_{j}
\end{align}
$\blacksquare$

\begin{statement}[Optimality condition for mode probabilities]
\label{statement:pcondition}
The following optimality condition 
\begin{equation}
\frac{\partial}{\partial \myvect{p}}
    \mathcal{M}\mathcal{W}^2_2(\mu_{GMM(\myvect{\Theta}^*, \myvect{p})}, \rho_{GMM(\myvect{\Gamma}, \myvect{q})})|_{\myvect{p}=\myvect{p}^*} = 0
\end{equation}
implies that
\begin{equation}\label{eq:optimality_modeprobs}
    \sum_{j=1}^J v^*_{ij} ||\myvect{\Theta}^*_i-\myvect{\Gamma}_j||_2^2 = const, \;\;\; \forall i \in \{1,...,I\}.
\end{equation}
% \akbar{I couldn't find the above condition elsewhere, and it is not really essential in our argument either.}\\
\textbf{Proof}: At the optimal point $[\mymatrix{\Theta}^*, \myvect{p}^*]$, if two mode probabilities like $p^*_{i_1}$ and $p^*_{i_2}$ are changed by $+\epsilon$ and $-\epsilon$ (where $\epsilon$ is a small positive scalar), we have that
\begin{align}
    \frac{change\; in \; \mathcal{M}\mathcal{W}^2_2}{change \; in \; \myvect{p}}\\
   =\frac{1}{\epsilon}\Big[\sum_{j=1}^J ||\myvect{\Theta}^*_{i_1}-&\myvect{\Gamma}_j||_2^2 \; v^*_{i_1j} \; (+\epsilon)\Big] + 
   \frac{1}{\epsilon}\Big[\sum_{j=1}^J ||\myvect{\Theta}^*_{i_2}-\myvect{\Gamma}_j||_2^2 \; v^*_{i_{2}j} \; (-\epsilon)\Big]\\
   = \Big[\sum_{j=1}^J ||\myvect{\Theta}^*_{i_1}-&\myvect{\Gamma}_j||_2^2 \; v^*_{i_1j}\Big] - 
   \Big[
    \sum_{j=1}^J ||\myvect{\Theta}^*_{i_2}-\myvect{\Gamma}_j||_2^2 \; v^*_{i_{2}j}
   \Big] = 0
\end{align}
Therefore
\begin{equation}
    \sum_{j=1}^J ||\myvect{\Theta}^*_{i_1}-\myvect{\Gamma}_j||_2^2 \; v^*_{i_1j} = \sum_{j=1}^J ||\myvect{\Theta}^*_{i_2}-\myvect{\Gamma}_j||_2^2 \; v^*_{i_{2}j}\;\;\; \;\;\;\; \forall i_1, i_2 \in \{1,...,I\}, \;\;i_1\neq i_2.
\end{equation}
$\blacksquare$
\end{statement}

Now we are ready to state the training phase and testing phase of the learning algorithm.

\begin{statement}[Training phase]\label{statement:training_phase}
Given is a set of $N$ GMM-induced measures $\{\rho_{GMM(\mymatrix{\Gamma}, \myvect{q}^{(n)})}\}_{n=1}^N$ and their corresponding population descriptors $\{\myvect{y}_1, .., \myvect{y}_N\}$. Note that the matrix $\mymatrix{\Gamma}\in \mathbb{R}^{J\times D}$ are the GMM modes in the target distribution that carpet the feature space, and are kept fixed during trianing and testing, hence to superscript $n$ for $\mymatrix{\Gamma}$. For each dataset instance, the following vectors and matrices are obtained
\begin{enumerate}\itemsep0pt
\item The parameters of the closest GMM to $\rho_{GMM(\mymatrix{\Gamma}, \myvect{q}^{(n)})}$ and with $I$ modes:
\begin{equation}
     \mymatrix{\Theta}^{*(n)}, \myvect{p}^{*(n)} = \underset{\substack{\mymatrix{\Theta}\in\mathbb{R}^{I\times D} \\ \myvect{p}\in \mathbb{R}^I }}{argmin}
    \;\; \mathcal{M}\mathcal{W}^2_2(\mu_{GMM(\myvect{\Theta}^, \myvect{p})}, \rho_{GMM(\myvect{\Gamma}, \myvect{q}^{(n)})})
\end{equation}
\item Having obtained $\mymatrix{\Theta}^{*(n)}$ and $\myvect{p}^{*(n)}$, solve the dual form of Eq. \ref{eq:kantorovich_dualform} for the distributons $\mu_{GMM(\mymatrix{\Theta}^{*(n)}, \myvect{p}^{*(n)})}$ and $\rho_{GMM(\myvect{\Gamma}^{(n)}, \myvect{q}^{(n)})}$. Save the first $I$ elements of the dual solution $[z^{*(n)}_1, ..., z^{*(n)}_I]$.
\end{enumerate}
Now having obtained $\{ \mymatrix{\Theta}^{*(n)}, \myvect{p}^{*(n)}, [z^{*(n)}_1, ..., z^{*(n)}_I]\}_{n=1}^N$, the following predictors are trained

\begin{enumerate}\itemsep0pt
    \item $\mymatrix{\Theta}^{*(n)}\approx h_{\mymatrix{\Theta}}(\myvect{y}_n)$, where $h_{\mymatrix{\Theta}}:\mathbb{R}^{dim(\myvect{y})}\rightarrow \mathbb{R}^{I\times D}$.
    \item $\myvect{p}^{*(n)}\approx h_{\myvect{p}}(\myvect{y}_n)$, where $h_{\myvect{p}}:\mathbb{R}^{dim(\myvect{y})}\rightarrow \mathbb{S}^{I-1}$.
    \item $[z^{*(n)}_1, ..., z^{*(n)}_I] \approx h_z(\myvect{y}_n)$, where $h_{z}:\mathbb{R}^{dim(\myvect{y})} \rightarrow \mathbb{R}^I$
\end{enumerate}
\end{statement}

\begin{statement}[Testing phase]\label{statement:testing_phase}
In the testing phase, a population descriptor $\myvect{y}_{test}$ is given. Firstly, the following vectors/matrices are predicted:

\begin{enumerate}\itemsep0pt
\item $\mymatrix{\Theta}^{(test)}\approx h_{\mymatrix{\Theta}}(\myvect{y}_{test})$.
\item $\myvect{p}^{(test)}\approx h_{\myvect{p}}(\myvect{y}_{test})$
\item $[z^{(test)}_1, ..., z^{(test)}_I] \approx h_z(\myvect{y}_{test})$
\end{enumerate}
Now having obtained $\big[\mymatrix{\Theta}^{(test)}, \myvect{p}^{(test)}, [z^{(test)}_1, ..., z^{(test)}_I]\big]$, according to Statement \ref{statement:Vstargt0_unique} $\mymatrix{V}^{test}_{\ge0}$ is uniquely identified and according to Statement \ref{statement:subsetsum:almostsurely} almost surely  has exactly $I+J-1$ non-zero elements. In other words, at this point we are left with $I+J-1$ variables (i.e. $v^{test}_{ij}$-s) that we want to identify as closely as possible. For these $I+J-1$ variables, we have the following constraints
\begin{enumerate}\itemsep0pt
\item Each barycentre condition in Eq. \ref{eq:barycentercondition} provides $D$ linear constraints on non-zero $v^{test}_{ij}$-s, resulting in $I\times D$ linear constraints. Note that the matrix $\mymatrix{\Gamma}\in \mathbb{R}^{J\times D}$ are the GMM modes in the target distribution that carpet the feature space, and are kept fixed throughout.
\item Each constraint of Eq. \ref{eq:optimality_modeprobs} provides a linear constraint, resulting in $I$ more linear constraints. We add the constant scalar in Eq. \ref{eq:optimality_modeprobs} to our variable list, so we have $I+J$ variables to identify.
\end{enumerate}

In sum, we have $I+J$ variables, $ID+I$ constraints, and hence $J - ID$ degrees of freedom.
Having estimated $\mymatrix{\Theta}^{(test)}$ and the flow $\mymatrix{V}^{(test)}$, we can apply the latter on the former to arrive at an estimate for $\rho^{(test)}$ 
\end{statement}

% \akbar{Some notes about the testing phase
% \begin{itemize}
%     \item[-] The system of linear equations is underdetermined if $I\times D < J$, but there are methods that pick, e.g., the solution with minimum 2-norm.
%     \item[-] In general for an underdetermined linear system if a single solution is picked in that subspace, there is no uppper bound on the error. \textbf{**** BUT, here we know that all the variables involved are bounded (e.g. $v_{ij}$-s are between 0 and 1.0) $\rightarrow$ there might be an upper bound for the error as that subspace gets smaller*****}
% \end{itemize}
% }

\begin{statement}\label{statement:proof_illdefined}
If $I=1$ the dual problem of Eq. \ref{eq:kantorovich_dualform} is ill-defined.
% \akbar{According to google search (AI mode) the statement is definitely true, but I've written the proof below myself.}
\textbf{Proof}: If $I=1$, according to Eq. \ref{eq:kantrovich_dual_complementary}, we have that
\begin{equation}
    z^*_{j+I} = \||\myvect{\Theta}_1-\myvect{\Gamma}_j||_2^2 - \frac{z^*_1}{\cancelto{1}{p_1}}.
\end{equation}
If we substitute this in Eq. \ref{eq:kantorovich_dualform}
\begin{align}
    \underset{\myvect{z}\in \mathbb{R}^{1 + J}}{sup} \; &z_1 + \sum_{j=1}^J q_{j} \; z_{j+I} =\\
    \underset{\myvect{z}\in \mathbb{R}^{1 + J}}{sup} \; &z_1 + \sum_{j=1}^J q_{j} \; \Big[
    \||\myvect{\Theta}_1-\myvect{\Gamma}_j||_2^2 - \frac{z_1}{\cancelto{1}{p_1}}
    \Big] =\\
    \underset{\myvect{z}\in \mathbb{R}^{1 + J}}{sup} \; &z_1 -z_1\cancelto{1}{\sum_{j=1}^Jq_j} + \sum_{j=1}^J q_{j} \; \Big[
    \||\myvect{\Theta}_1-\myvect{\Gamma}_j||_2^2 
    \Big]
\end{align}
In the last equation $z_1$ is cancelled out from the objective.
$\blacksquare$
\end{statement}

Essentially, according to Statement~\ref{statement:proof_illdefined} the described training and testing procedure fail, and this discourages the use of uni-modal Gaussian distribution as the base distribution.

\section{Letters}

\subsection{Dataset details}
\subsection{Synthetic Letters Dataset (Letter--Rotation Benchmark)}
\label{app:letters}

\paragraph{Data generation.}
We construct a synthetic benchmark by rendering uppercase letters as binary silhouettes and sampling point clouds from the foreground pixels. Each condition corresponds to a (letter, rotation) pair. For a given letter, we generate $R=20$ rotations uniformly spaced in $[0,2\pi]$. For each condition, we create multiple replicas by rendering the same silhouette with different RGB colours; these replicas are used only to provide multiple samples from the same conditional distribution and are not included in the conditioning vector.

\paragraph{Point-cloud representation.}
Each sample is a point cloud in $\mathbb{R}^{D}$ with $D=5$, consisting of 2D coordinates and RGB channels:
$
\myvect{x} = (x, y, r, g, b).
$
We extract foreground-pixel coordinates, normalize them to $[-1,1]^2$, rescale to correct aspect ratio using the glyph’s bounding box, optionally add Gaussian coordinate noise (set to $0$ in our experiments), and finally rotate the point cloud by the condition’s angle.

\paragraph{Colour replicas.}
For each (letter, rotation) condition, we generate $C$ colour replicas to better approximate the underlying conditional distribution. In our experiments we use $C=100$. To ensure that colour does not dominate the evaluation metrics, we restrict the colour variation to a narrow range by sampling only the red channel $r \sim \mathrm{Unif}(r_{\min}, r_{\max})$ with $(r_{\min}, r_{\max})=(0.5,0.6)$ and setting $g=b=0$; thus colour provides mild diversity without strongly affecting distance metrics.

\paragraph{Condition descriptor.}
The conditioning vector $\myvect{y}$ concatenates (i) a one-hot encoding of the letter identity and (ii) a continuous rotation scalar normalized to $[0,1]$:
\[
\myvect{y} = [\mathrm{onehot}(\text{letter});\;\theta/(2\pi)].
\]
Colour is not included in $\myvect{y}$.

\paragraph{Source distribution.}
For conditional flow-matching training we sample the source point cloud from an isotropic Gaussian. In the our setting, each source point cloud uses a shared RGB value across all points (sampled uniformly within the same restricted colour range) while the 2D coordinates are sampled from $\mathcal{N}(0, I)$, matching the data format of the target clouds.

\paragraph{Train/validation/test splits (rotation interpolation).}
We study rotation interpolation by splitting the rotation indices: training uses even-indexed rotations, while validation and test use odd-indexed rotations. All letters appear in training (but only at even rotations). The validation set uses a designated validation letter (\texttt{S} in the main paper), evaluated on odd rotations. Test sets are formed from unseen test letters (\texttt{W} and \texttt{Y} in the main paper), also evaluated on odd rotations. Additional results with alternative validation/test letters are reported in the Table \ref{tab:synthetic_combined_reformatted_app}.

\paragraph{Subsampling for OT coupling.}
Each rendered silhouette contains many foreground pixels; to make mini-batch OT pairing tractable, we subsample a fixed number of points per cloud. We use $n=100$ points per cloud because the discrete OT coupling has super-cubic complexity in $n$ in our implementation, making larger clouds prohibitively expensive. This subsampling is applied consistently across all methods.

\paragraph{Key hyperparameters.}
Unless otherwise stated, we use: batch size $10$, point subsampling $n=100$, number of letters $4$, number of rotations $R=20$, number of colour replicas per condition $C=100$, $(r_{\min},r_{\max})=(0.5,0.6)$, and zero coordinate noise.

\subsection{SP-FM implementation}

\subsection{SP-FM / PertFlow on Colored Letter Point Clouds}
\label{sec:spfm_pertflow}

\paragraph{Data representation and conditions.}
Each sample is a point cloud of $N$ points, where each point carries 2D geometry and RGB color:
\[
x \;=\; (x,y,r,g,b)\in\mathbb{R}^5, 
\qquad 
\mathbf{X}\in\mathbb{R}^{N\times 5}.
\]
We condition generation on the letter identity and rotation. The condition vector is
\[
\mathbf{y} = [\text{one-hot}(\text{letter}),\, \rho]\in\mathbb{R}^{C},
\]
where $\rho\in[0,1]$ is the normalized rotation angle (mapped to $[0,2\pi]$ when decoding for visualization).

\paragraph{Dual-flow architecture (geometry vs.\ color).}
We use a dual-MLP conditional flow architecture:
(i) a geometry flow $v_{\text{xy}}$ acting on $(x,y)\in\mathbb{R}^2$ \emph{per point}, and
(ii) a color flow $v_{\text{rgb}}$ acting on an RGB vector $(r,g,b)\in\mathbb{R}^3$ \emph{per cloud}.
Concretely, we decompose each cloud as
\[
\mathbf{X}_{\text{xy}} \in \mathbb{R}^{N\times 2},
\qquad
\mathbf{c} \in \mathbb{R}^{1\times 3},
\]
where $\mathbf{c}$ denotes the (shared) RGB color of the cloud (in our setup, all points in a cloud share the same color).
Both velocity fields are implemented as MLPs that take time $t$ and the concatenation of state and condition:
\[
v_{\text{xy}}(t,\mathbf{x}_{\text{xy}};\mathbf{y}) 
\;=\; \mathrm{MLP}_{\text{xy}}\!\big(t,\,[\mathbf{x}_{\text{xy}},\mathbf{y}]\big),
\qquad
v_{\text{rgb}}(t,\mathbf{c};\mathbf{y}) 
\;=\; \mathrm{MLP}_{\text{rgb}}\!\big(t,\,[\mathbf{c},\mathbf{y}]\big).
\]

\paragraph{Conditional base distribution with random per-condition mode initialization.}
To sample the initial state $\mathbf{X}_0$ we use a condition-dependent mixture base.
In the \texttt{use\_kmeans=True} setting, mixture \emph{centers} are initialized by \textbf{randomly sampling} $K$ training point clouds for each condition $\mathbf{y}$ (i.e., for each letter-rotation pair), rather than running K-means. These sampled clouds serve as fixed centers, while the mixture weights are predicted by a small network $h_p(\mathbf{y})$.
At sampling time we draw a (soft) mixture over centers via Gumbel--Softmax (temperature $\tau$) and add Gaussian perturbation with variance $\sigma^2$ to obtain $\mathbf{X}_{0,\text{xy}}$.

For color, we initialize a single RGB value per cloud by sampling uniformly from a predefined range $[a,b]\subset[0,1]$ and broadcasting it to all points:
\[
\mathbf{c}_0 \sim \mathcal{U}([a,b]^3), 
\qquad 
\mathbf{X}_{0,\text{rgb}} = \mathbf{1}_N \mathbf{c}_0,
\]
and finally concatenate geometry and color to obtain $\mathbf{X}_0=[\mathbf{X}_{0,\text{xy}},\mathbf{X}_{0,\text{rgb}}]\in\mathbb{R}^{N\times 5}$.

\paragraph{OT pairing.}
Before computing training losses, we align points between the sampled source cloud $\mathbf{X}_0$ and the target cloud $\mathbf{X}_1$ using point-wise optimal transport (OT) within each sample, which reduces sensitivity to arbitrary point ordering.

\paragraph{Flow matching objective.}
Given an OT-aligned pair $(\mathbf{X}_0,\mathbf{X}_1)$ under condition $\mathbf{y}$, we sample $t\sim\mathcal{U}[0,1]$ and define
\[
\mathbf{X}_t = (1-t)\mathbf{X}_0 + t\mathbf{X}_1,
\qquad
\mathbf{U} = \mathbf{X}_1-\mathbf{X}_0.
\]
We use the standard flow-matching regression form
\[
\mathcal{L}_{\mathrm{FM}}
=
\mathbb{E}\Big[
\|v(t,\mathbf{X}_t;\mathbf{y})\|_2^2
-
2\langle v(t,\mathbf{X}_t;\mathbf{y}),\, \mathbf{U}\rangle
\Big].
\]
With the dual architecture, we compute this loss separately for geometry and color and sum them:
\[
\mathcal{L}_{\mathrm{FM}}
=
\mathcal{L}_{\mathrm{FM}}^{\text{xy}}
+
\mathcal{L}_{\mathrm{FM}}^{\text{rgb}},
\]
where $\mathcal{L}_{\mathrm{FM}}^{\text{xy}}$ is evaluated over all $N$ points, and $\mathcal{L}_{\mathrm{FM}}^{\text{rgb}}$ is evaluated on the single RGB vector per cloud.

\paragraph{Geodesic-length loss (default endpoint loss).}
In addition, we employ a shortest-path regularizer using the \textbf{default endpoint loss} (as in our configuration):
\[
\mathcal{L}_{\mathrm{geo}}
=
\mathbb{E}\big[\|\mathbf{X}_1-\mathbf{X}_0\|_2^2\big].
\]
In the dual setting, this decomposes as
\[
\mathcal{L}_{\mathrm{geo}}
=
\mathcal{L}_{\mathrm{geo}}^{\text{xy}}
+
\mathcal{L}_{\mathrm{geo}}^{\text{rgb}},
\qquad
\mathcal{L}_{\mathrm{geo}}^{\text{xy}}=\mathbb{E}\big[\|\mathbf{X}_{1,\text{xy}}-\mathbf{X}_{0,\text{xy}}\|_2^2\big],
\quad
\mathcal{L}_{\mathrm{geo}}^{\text{rgb}}=\mathbb{E}\big[\|\mathbf{c}_{1}-\mathbf{c}_{0}\|_2^2\big].
\]

\paragraph{Overall training objective.}
We optimize the combined objective
\[
\mathcal{L}
=
\mathcal{L}_{\mathrm{FM}}
+
\lambda\,\mathcal{L}_{\mathrm{geo}},
\]
with $\lambda$ controlling the strength of the geodesic-length regularization.

\subsection{GMM implementation}

\begin{table}[t]
\centering
\caption{Results on two synthetic experiments based on the letter~S and letter~H silhouettes. 
Each experiment assesses a model’s ability to generalize to rotations of the corresponding shape 
that are not observed during training.}
\label{tab:synthetic_combined_reformatted_app}
\resizebox{\linewidth}{!}{
\begin{tabular}{llcccccc}
\toprule
\multirow{2}{*}{Dataset} & \multirow{2}{*}{Model} 
& \multicolumn{3}{c}{Generated vs. Target} 
& \multicolumn{3}{c}{Source vs. Target} \\
\cmidrule(lr){3-5} \cmidrule(lr){6-8}
& & MMD $\downarrow$ & W1 $\downarrow$ & W2 $\downarrow$
  & MMD $\downarrow$ & W1 $\downarrow$ & W2 $\downarrow$ \\
\midrule

\multirow{2}{*}{Letter H (Validation)}
  & CFM
    & 0.02021 $\pm$ 0.00115 & 0.7044 $\pm$ 0.0222 & 0.8626 $\pm$ 0.0229
    & 0.09731 $\pm$ 0.00101 & 1.2101 $\pm$ 0.0056 & 1.3342 $\pm$ 0.0066 \\
  & SP-FM
    & \textbf{0.01387 $\pm$ 0.00092} & \textbf{0.5939 $\pm$ 0.0190} & \textbf{0.7372 $\pm$ 0.0198}
    & \textbf{0.02773 $\pm$ 0.00042} & \textbf{0.7631 $\pm$ 0.0053} & \textbf{0.9048 $\pm$ 0.0047} \\
\midrule
\multirow{2}{*}{Letter W (Test)}
  & CFM
    & 0.01778 $\pm$ 0.00101 & 0.6118 $\pm$ 0.0187 & 0.7219 $\pm$ 0.0203
    & 0.06088 $\pm$ 0.00049 & 0.8972 $\pm$ 0.0062 & 1.0151 $\pm$ 0.0071 \\
  & SP-FM
    & \textbf{0.01352 $\pm$ 0.00143} & \textbf{0.5429 $\pm$ 0.0196} & \textbf{0.6441 $\pm$ 0.0167}
    & \textbf{0.01872 $\pm$ 0.00062} & \textbf{0.6140 $\pm$ 0.0091} & \textbf{0.7173 $\pm$ 0.0099} \\

\multirow{2}{*}{Letter Y (Test)}
  & CFM
    & 0.02648 $\pm$ 0.00230 & 0.6669 $\pm$ 0.0295 & 0.7779 $\pm$ 0.0390
    & 0.06020 $\pm$ 0.00128 & 1.0614 $\pm$ 0.0048 & 1.2407 $\pm$ 0.0055 \\
  & SP-FM
    & \textbf{0.01735 $\pm$ 0.00167} & \textbf{0.5406 $\pm$ 0.0259} & \textbf{0.6358 $\pm$ 0.0295}
    & \textbf{0.04134 $\pm$ 0.00221} & \textbf{0.9525 $\pm$ 0.0399} & \textbf{1.1096 $\pm$ 0.0448} \\
\midrule

\end{tabular}%
}
\end{table}

% \section{Model Implementation and Hyperparameters}

\end{document}